\documentclass[journal,onecolumn]{IEEEtran}

\usepackage[vlined,ruled,linesnumbered]{algorithm2e}
\SetKwInOut{Input}{input}
\SetKwInOut{Output}{output}
\SetKwComment{Comment}{}{}

\SetCommentSty{mycmtsty}

\usepackage{setspace}

\usepackage{amsthm}

\usepackage{amsmath}
\usepackage{amsfonts}
\usepackage{amssymb}
\usepackage{multirow}
\usepackage{stmaryrd}
\usepackage{graphicx}
\usepackage{booktabs}

\usepackage{pdflscape}
\usepackage[T1]{fontenc}

\usepackage{hyperref}

\newcommand{\etal}{\textit{et al.\ }}
\newcommand{\Best}[1]{\textbf{\textcolor{black}{#1}}}
\newcommand{\B}[1]{\Best{#1}}
\newcommand{\SB}[1]{\textbf{\color{brown}{#1}}}

\newcommand{\ma}[1]{\boldsymbol{#1}}

\usepackage{tikz}
\usetikzlibrary{positioning}
\usetikzlibrary{shapes.geometric, arrows}

\tikzstyle{etape} = [rectangle, rounded corners, minimum width=3cm, minimum height=1cm, text width=3cm, text centered, draw=black]
\tikzstyle{ghost} = []
\tikzstyle{arrow} = [thick,->,>=stealth]

\usepackage{fancyvrb}
\VerbatimFootnotes %

\begin{document}
\title{Bare Demo of IEEEtran.cls\\ for IEEE Journals}

\author{Michael~Shell,~\IEEEmembership{Member,~IEEE,}
        John~Doe,~\IEEEmembership{Fellow,~OSA,}
        and~Jane~Doe,~\IEEEmembership{Life~Fellow,~IEEE}%
\thanks{M. Shell was with the Department
of Electrical and Computer Engineering, Georgia Institute of Technology, Atlanta,
GA, 30332 USA e-mail: (see http://www.michaelshell.org/contact.html).}%
\thanks{J. Doe and J. Doe are with Anonymous University.}%
\thanks{Manuscript received April 19, 2005; revised August 26, 2015.}}

\title{Implementation of the VBM3D Video Denoising Method and Some Variants}

\author{Thibaud Ehret, Pablo Arias%
\thanks{Both authors are with CMLA, CNRS, ENS Paris-Saclay, Universit{\'e}
Paris-Saclay. e-mail: \texttt{thibaud.ehret@cmla.ens-cachan.fr}}%
\thanks{Work partly financed by IDEX Paris-Saclay IDI 2016,
ANR-11-IDEX-0003-02, Office of Naval research grant N00014-17-1-2552, DGA
Astrid project \guillemotleft filmer la Terre\guillemotright~n$^o$
ANR-17-ASTR-0013-01, MENRT.}
}

\markboth{Preprint, January 2020.}%
{Ehret, Arias: Implementation of the VBM3D Video Denoising Method and Some Variants}

\maketitle

\begin{abstract}
	VBM3D is an extension to video of the well known image denoising algorithm BM3D, which takes advantage of the sparse representation of stacks of similar patches in a transform domain. The extension is rather straightforward: the similar 2D patches are taken from a spatio-temporal neighborhood which includes neighboring frames. In spite of its simplicity, the algorithm offers a good trade-off between denoising performance and computational complexity. In this work we revisit this method,  providing an open-source C++ implementation reproducing the results. A detailed description is given and the choice of parameters is thoroughly discussed. Furthermore, we discuss several extensions of the original algorithm: (1) a multi-scale implementation, (2) the use of 3D patches, (3) the use of optical flow to guide the patch search. These extensions allow to obtain results which are competitive with even the most recent state of the art.
\end{abstract}

\section{Introduction}

 VBM3D was proposed by \cite{Dabov2007v} as an adaptation to video denoising of BM3D, the  successful image denoising algorithm \cite{dabov2007image}. The algorithme is designed for  additive white Gaussian noise with zero mean and standard deviation $\sigma$, \textit{i.e.} 
\[v(\rho) = u(\rho) + n(\rho), \quad n(\rho) \sim \mathcal N(0,\sigma^2),\]
where $\rho$ is a position in the video domain (a pixel), $v$ is the noisy video and $u$ the unknown clean video.
The denoising principle of VBM3D (and BM3D) is based on the redundancy of similar patches. 
Groups of similar 2D patches are assembled in a 3D stack of patches. A separable 3D transform is applied to this stack. The stack is denoised by applying a shrinkage operator to the coefficients in the transformed domain. 
The algorithm follows four basic steps:
\begin{enumerate}
	\item Search for similar patches in the sequence, grouping them in a 3D stacks,
	\item Apply a 3D linear domain transform to the 3D block,
	\item Shrink the transformed coefficients,
	\item Apply the inverse transform,
	\item Aggregate the resulting patches in the video.
\end{enumerate}
The underlying idea here is that, due to the high redundancy of a stack of similar patches, the energy will be concentrated in a few coefficients of the transform, while the noise is spread evenly among all coefficients. This allows to jointly denoise the patches of each stack. 
The reconstruction of the estimated video is obtained by aggregating for each pixel the estimated patches that  contain  it. 
This principle is applied twice. The first time the patches are denoised using a hard threshold in the transformed domain. In the second iteration a Wiener filter is used, with Wiener coefficients computed using the output of the first step as oracle.
There exist other adaptations to video and 3D images of this framework: BM4D \cite{Maggioni2013} and VBM4D \cite{Maggioni2012}. These methods stack similar 3D spatio-temporal patches in a 4D stack. 
The main difference between them is that VBM4D uses motion compensated patches whereas BM4D aims at denoising volumetric images (such as those appearing in medical imaging), where the 3rd dimension is  just another spatial dimension. In \cite{arias-18-patch-models} however, it was shown that even  non-motion compensated 3D patches  provide a very good denoising performance.

In this article, we revisit the VBM3D method, providing an open source implementation and we discuss some variants introduced in \cite{arias-18-patch-models}, such as 3D patches (as in BM4D \cite{Maggioni2013}), optical flow guided patch search, and a multiscale implementation.
In the next section, the algorithm itself is reviewed. %
In the following sections we consider three possible extensions: multiscale, spatio-temporal patches and motion compensated patch search using an optical flow.
Finally the performance of the algorithm is compared against other state of the art algorithms.

\paragraph{Notation }
 
The noisy input video denoted $v$  consists of $f+1$ frames written $v_i$ for $i$ between $0$ and $f$. The clean unknown video is denoted by $u$. A patch is a small rectangular piece of the video, for example of size $8\times 8\times 3$ ($8$ pixels width and height, 3 pixels in the temporal dimension). Patches are represented by bold lowercase letters, \textit{e.g.} $\ma p$. If we need to emphasize the location of the patch we write $\ma p(x)$, where $x$ represents the location on the video of the top-left-front pixel of the patch.

For the parameters of the method we will use the same notation as in \cite{Dabov2007v}. The different parameters for the patch search are the number $N$ of patches to return, the number of temporal frames $N_f$ that are being searched, the search region for the reference frame $N_s$, the search region for the other frames $N_{pr}$, the maximum number $N_b$ of patches to compute for each local search, a correcting factor $d$ and a maximum distance threshold $\tau$. The threshold parameter in the first step is $\lambda_{3D}$. We also write $\otimes$ for the element-wise product.
Similar parameters will be used for  the hard thresholding first step and for the Wiener filtering second step. Because they can have different values for the different steps, they'll be differentiated using the subscript $hard$ and $wien$ respectively. 

\section{VBM3D :  the algorithm}

\begin{figure}[ht]
	\centering
	\includegraphics[width=\linewidth]{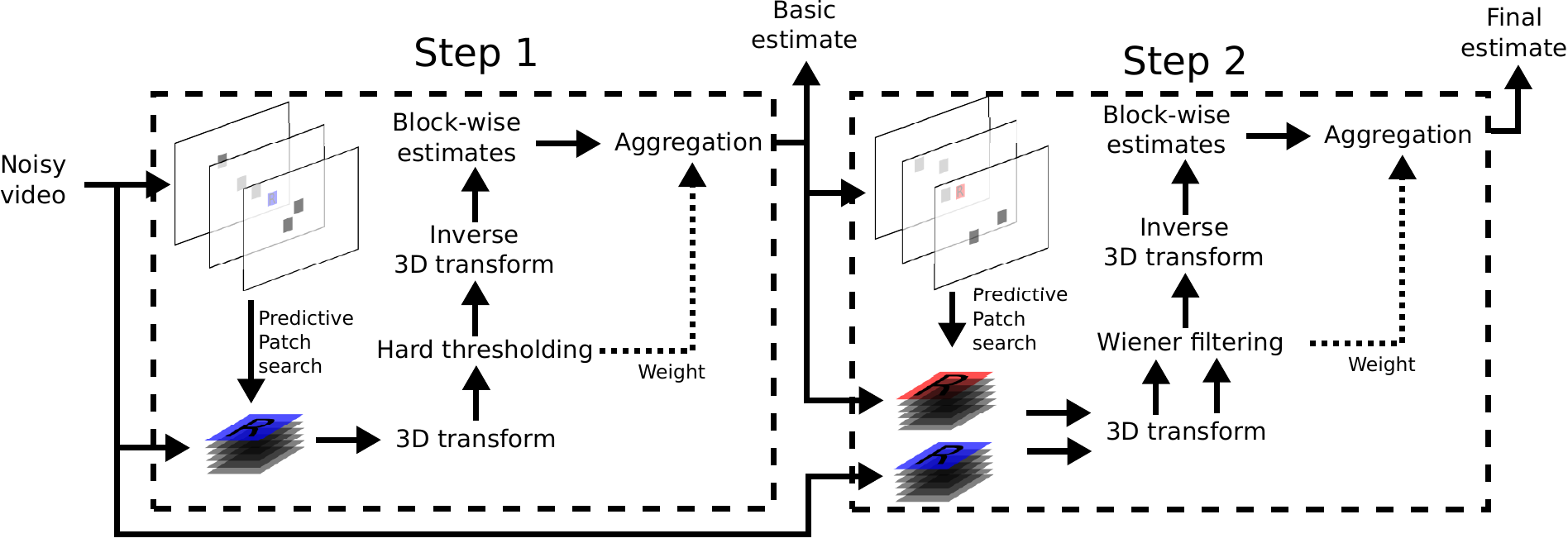}
	\caption{Scheme of the core of the VBM3D algorithm.}
	\label{fig:core}
\end{figure}

 VBM3D performs two steps, as its image counterpart BM3D. The first \emph{hard-thresholding} step computes a a basic estimate which is refined in the \emph{Wiener filtering} step, yielding the final estimate. Figure \ref{fig:core} represents the global structure of the algorithm. 
The pseudocode of the core of the algorithm is presented in Algorithm \ref{alg:vbm3d}.

\subsection{The patch search}
\label{sec:patch_search}

\begin{figure}[ht]
	\centering
	\includegraphics[width=\linewidth]{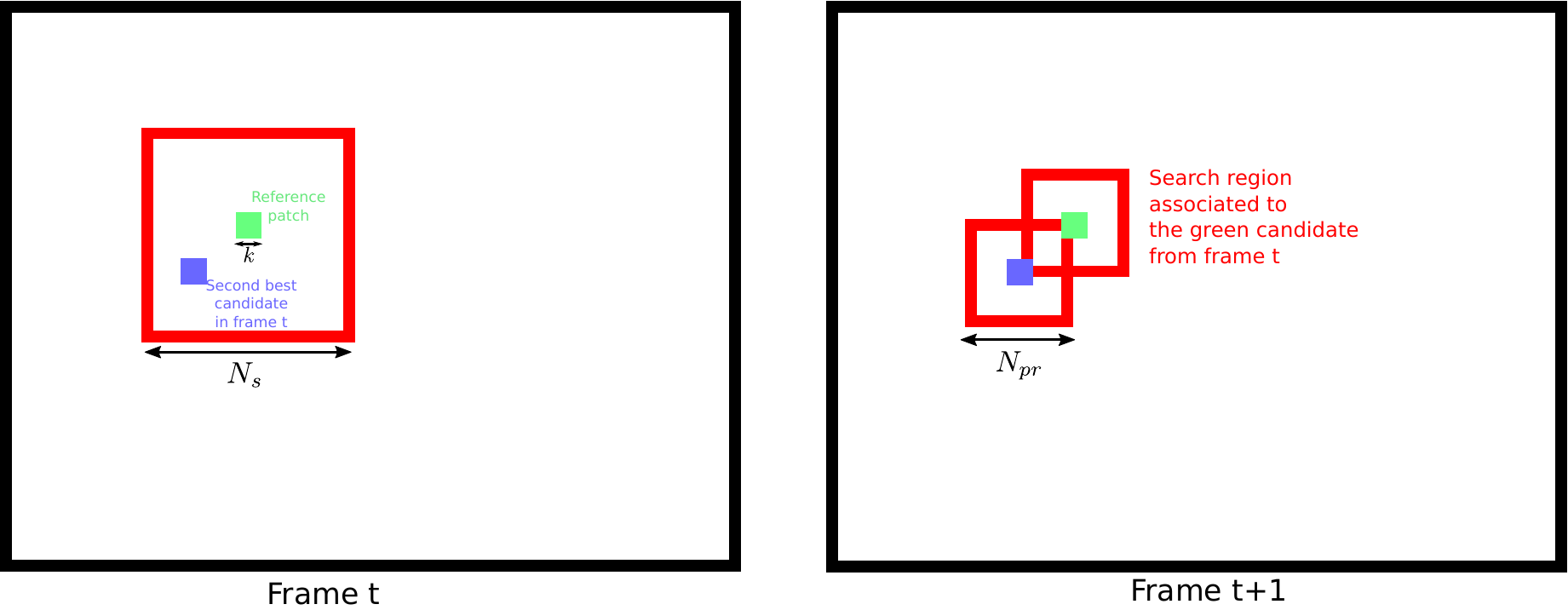}
	\caption{The patch search starts with a local spatial search in the current frame of size $N_s \times N_s$. Only the $N_b$ best candidates are then used in the following frame to predict where to search. This leads to $N_b$ local spatial searches (each centered at the candidates from the previous frame), of size $N_{pr}\times N_{pr}$. In practice, $N_s = 7$, $N_{pr} = 5$ and $N_b = 2$.}
	\label{fig:patch_search}
\end{figure}

The groups of similar patches are built by selecting a reference patch and searching around it for its nearest neighbors.
This patch search is the main difference between BM3D and VBM3D. The image version of the algorithm searches a square 2D window centered at the reference patch. While the same could be done in a space-time volume for videos, Dabov et al. \cite{Dabov2007v} propose a \emph{predictive search} heuristic to reduce the size of the search window. The idea behind the proposed search is to track patches in the video. 
Let $\ma p$ and $\ma q$ be two patches at positions $(x,y,t)$ and $(x',y',t')$ respectively. The distance between $\ma p$ and $\ma q$ used to find the nearest neighbors for a given regularizing factor $d$ is
\begin{equation}
	\label{eq:distance}
	d(\ma p, \ma q) = 
	\begin{cases}
		\|\ma p - \ma q\|^2_2 - d \text{ if } x = x' \text{ and } y = y'\\
		\|\ma p - \ma q\|^2_2 \text{ otherwise}
	\end{cases}.
\end{equation} 
This distance regularizes the patch trajectories by favoring non moving patches.
Suppose that the reference patch is $\ma p$  located at $(x,y,t)$. The goal of the proposed search is to find (at most) $N$ patches in a window of $2N_f+1$ frames around frame $t$. The steps are the following:
\begin{enumerate}
	\item Find the $N_b$ nearest neighbors to $\ma p$ in frame $t$ in a square search region of size $N_s\times N_s$ centered at the location of the reference patch $(x,y,t)$. Let $L_t$ be the set of found patches.
	\item Find the $N_b$ nearest neighbors to $\ma p$ in frames $t' = t+1,...,t+N_f$. The search region at $t'$ is the union of $N_{pr}\times N_{pr}$ square regions centered at positions of the candidates in $L_{t'-1}$, where $N_{pr} < N_s$. This is depicted in Figure \ref{fig:patch_search}.
	\item Similarly, find $N_b$ nearest neighbors to $\ma p$ in frames $t' = t-1, t-2, ..., t-N_f$. This time the search region is the union of squares centered at the positions of the patches in $L_{t'+1}$.
	\item Remove the candidates with distance \eqref{eq:distance} larger than $\tau$ from
	\[L = \bigcup_{t' = t-N_f}^{t+N_f} L_{t'},\] 
	the set combining all the candidates computed in each frame.
	\item Keep the best $N$ candidates from $L$.
	\item Because  the  next steps of the algorithm (in particular the domain transforms) require a number of patches that is a power of $2$, only the largest power of $2$ smaller or equal to $N$ is kept. 	
\end{enumerate}
These steps are summarized in Algorithms \ref{alg:vbm3d_search} and \ref{alg:local_search}.

\begin{algorithm}
	\caption{\texttt{compute\_similar\_patches}: VBM3D patch search}
	\label{alg:vbm3d_search}
	\setstretch{1.4}
	\DontPrintSemicolon
	\Input{Noisy video $v = (v_0, \dots, v_f)$, a reference patch $\ma p$, the number of patches to return $N$, a number of temporal frames $N_f$, a search region for the reference frame $N_s$, a search region for the other frames $N_{pr}$, the maximum number of patches to compute with each local search $N_b$, the size of the patch $k$, a correcting factor $d$, a maximum distance threshold $\tau$}
	\Output{$L$ the list of the $N$ patches closest to $\ma p$ and their distance}
	$t \leftarrow$ frame at which $\ma p$ is located\;
    $L_{t} \leftarrow \texttt{local\_search}(\ma p, \ma p, N_s, k, N_b, d, u_t)$ \tcp{Centered on $\ma p$ using $\ma p$ as a reference} 
	\tcp{Search in the $N_f$ following frames }
	\For{$t_f = (t+1)$ to $\min(t+N_f, f-1)$}{
		\For{$\ma q \in L_{t_f - 1}$}{
			$L_{t_f} \leftarrow \texttt{local\_search}(\ma q, \ma p, N_{pr}, k, N_b, d, u_{t_f})$ \tcp{Centered on $\ma q$ using $\ma p$ as a reference} 
		}
	}
	\tcp{Search in the $N_f$ previous frames}
	\For{$t_p = t-1$ to $\max(t-N_f, 0)$}{
		\For{$\ma q \in L_{t_p + 1}$}{
			$L_{t_p} \leftarrow \texttt{local\_search}(\ma q, \ma p, N_{pr}, k, N_b, d, u_{t_p})$ \tcp{Centered on $\ma q$ using $\ma p$ as a reference} 
		}
	}
	$L \leftarrow \bigcup_{i \in \llbracket t-N_f; t+N_f \rrbracket} L_i$\;
	$L \leftarrow $ elements from $L$ with distance smaller than $\tau$\;
	\tcp{the 3D transform requires a number of patches which is a power of $2$}
	$N_l \leftarrow $closest power of $2$ small or equal than $min(size of L, N)$\;
	\If{$L$ has more than $N_l$ elements}{
		$L \leftarrow N_l$ best candidate from $L$\;
	}
	\Return $L$\;
\end{algorithm}

\begin{algorithm}
	\caption{\texttt{local\_search}: Local search}
	\label{alg:local_search}
	\setstretch{1.4}
	\DontPrintSemicolon
	\Input{A patch $\ma p$ center of the search region, a reference patch $\ma p'$, the search region size $s$, the size of the patch $k$, the number of patches to return $N_b$, a correcting factor $d$, a frame $u$}
	\Output{$L$ the list of the patches closest to $\ma p'$ in the region described by $\ma p$ and $s$ and their distance}
	\tcp{Compute the distances for all the patches in the local region}
	\For{each patch $\ma q$ of size $k \times k$ in the spatial region of size $s \times s$ centered on $\ma p$}{
		\eIf{$\ma q$ is at the same spatial position than $\ma p$}{
			Add $(\|\ma q - \ma p'\|^2_2 - d, \ma q)$ to $L$\;    
		}{
			Add $(\|\ma q - \ma p'\|^2_2, \ma q)$ to $L$\;    
		}
	}
	Sort $L$ according to the value of the distance\;
	Keep the $N_b$ best elements in $L$\;
	\Return $L$\;
\end{algorithm}

\subsection{Patch stack filtering: hard-threshold step}
\label{sec:first_step}

The first step processes the image by filtering stacks of similar patches and aggregating them in an output image. The reference patches for the stacks are all patches on a sub-grid of step $st_{hard}$. For each group there  is first an estimation, followed by an aggregation.

\paragraph{Estimation.}
For a given reference patch $\ma p$, 
we first find its similar patches $L$ as described in Section \ref{sec:patch_search}. These patches are stacked in a 3D volume $\mathcal P(\ma p)$, of size $k\times k\times N$. The first spatial slice of this stack is the reference patch, and the remaining ones are the similar patches in $L$ ordered by their distance to $\ma p$. A 3D domain transform is first applied to the group of patches followed by a thresholding of the resulting spectrum (excepting the DC component of every patch). The inverse 3D domain transform is then applied to the thresholded coefficients to obtain the estimation, \textit{i.e.}
\begin{equation}
	\widehat{\mathcal{P}(\ma p)} = T_{3D}^{-1}(HT_\lambda(T_{3D}(\mathcal{P}(\ma p))))
	\label{eq:hard_thresholding}
\end{equation}
where $HT$ is defined by
\begin{equation}
	HT_{\lambda \sigma}(\nu) = 
	\begin{cases}
		\nu \text{ if $\nu$ is a DC component or } |\nu| > \lambda\sigma\\
		0 \text{ otherwise.}
	\end{cases}
	\label{eq:HT}
\end{equation}
In practice the 3D transforms are chosen to be separable, consisting of a 2D spatial transform applied directly on the patches of $\mathcal{P}(\ma p)$, typically a bi-orthogonal wavelet transform, followed by an 1D transform along the third dimension of the stack, typically a Haar transform.
The pseudocode of the estimation for the first step is presented in Algorithm \ref{alg:ht}.

\paragraph{Aggregation.}
An output pixel is estimated several times, since it belongs to several patches and each patch can be estimated multiple times (once for each group it belongs to). To compute the output frame $\widehat u$ these estimates are  aggregated. For a pixel $\rho$ of the frame, this aggregation is performed as
\begin{equation}
    \widehat{u}(\rho) = \frac{\sum_{\text{patch $\ma p$ of $v$}} \sum_{\ma q \in \widehat{\mathcal{P}(\ma p)}} \omega_{\ma p}(\rho) \ma q(\rho)}{\sum_{\text{patch $\ma p$ of $v$}} \omega_{\ma p}(\rho)}	
	\label{eq:aggregation}
\end{equation}
where a patch $\ma p$ has a support of the size of the image and is zero everywhere except on the actual location of the patch.
The weights $\omega$ used during the aggregation are computed in part during the estimation. The part $w_{ht}$ computed during the estimation is $w_{ht}(\ma p) = \frac{1}{\sigma^2N_{hard}(\ma p)}$ where $N_{hard}(\ma p)$ corresponds to the number of coefficients that have not been thresholded during the estimation of $\widehat{\mathcal{P}(\ma p)}$. Each coefficient $w_{ht}(\ma p)$ is properly defined because $N_{hard}(\ma p)$ is always positive as the DC component is never thresholded. The rest of the weight comes from a 2D Kaiser window $K(\ma p)$ of size $k_{hard}\times k_{hard}$ (see Eq. \eqref{eq:kaiser}) located on the position of $\ma p$ applied to avoid boundary effects from the patch.
The Kaiser window of size $L_x \times L_y$ of parameter $\beta$ is 
\begin{align}
    K(\ma p)(x,y) = I_0\left(\beta \sqrt{1 - (2x/L_x)^2}\right) I_0\left(\beta \sqrt{1 - (2y/L_y)^2}\right) / I_0(\beta)^2\\ \text{ with } 0 \leqslant x \leqslant L_x, 0 \leqslant y \leqslant L_y \text{ and $I_0$ the zeroth-order modified Bessel function}
    \label{eq:kaiser}
\end{align}
Finally the (patch) weight is defined by $\omega_{\ma p} = w_{ht}(\ma p) K(\ma p)$. The pseudocode with the aggregation step can be found in Algorithm \ref{alg:vbm3d}.

\begin{algorithm}
	\caption{\texttt{ht\_filtering}: Hard thresholding}
	\label{alg:ht}
	\setstretch{1.4}
	\DontPrintSemicolon
	\Input{A group of similar patches $L$, a 3D transform $T_{3D}$, the noise variance $\sigma^2$}
	\Output{A list of filtered patches $\hat{L}$, the aggregation weight $\omega$}
	$L \leftarrow T_{3D}(L)$\;
	$n \leftarrow 0$\;
	\For{each patch $\ma p$ in $L$}{
		\For{each pixel $\rho$ of $\ma p$}{
			\eIf{$\ma p(\rho) > \lambda \sigma$ or $\rho$ is the DC component}{
				$n \leftarrow n + 1 $\;
				$\hat{\ma p}(\rho) \leftarrow \ma p(\rho)$\;
			}{
					$\hat{\ma p}(\rho) \leftarrow 0$\;
			}
		}
	} 
	$\omega= \frac{1}{\sigma^2n}$\;
	$\hat{L} \leftarrow T_{3D}^{-1}(L)$\;
	\Return $\hat{L}, \omega$\;
\end{algorithm}

\begin{algorithm}
	\caption{\texttt{wiener\_filtering}: Wiener thresholding}
	\label{alg:wt}
	\setstretch{1.4}
	\DontPrintSemicolon
	\Input{A group of similar patches $L$, a first estimate of $L$ called $L'$, a 3D transform $T_{3D}$, the noise variance $\sigma^2$}
	\Output{A list of filtered patches $\hat{L}$, the aggregation weight $\omega$}
	$L \leftarrow T_{3D}(L)$\;
	$\omega \leftarrow 0$\;
	\For{each patch $\ma p$ in $L$, $\ma p'$ the corresponding patch in $L'$}{
		\For{each pixel $\rho$ of $\ma p$}{
			$\alpha \leftarrow \frac{\ma p'(\rho)^2}{\ma p'(\rho)^2 + \sigma^2}$\;
			$\hat{\ma p}(\rho) \leftarrow \alpha \ma p(\rho)$\;
			$\omega \leftarrow \omega + \alpha^2$\;
		}
	} 
	$\omega \leftarrow \frac{1}{\sigma^2\omega}$\;
	$\hat{L} \leftarrow T_{3D}^{-1}(\hat{L})$\;
	\Return $\hat{L}$, $\omega$\;
\end{algorithm}

\subsection{Patch stack filtering: Wiener filtering step}
\label{sec:second_step}

The second step of the algorithm uses the basic estimate computed during the first step for the patch search, and to compute the coefficients of a Wiener shrinkage of the transformed coefficients.
The video is processed by building groups of patches around reference patches from a coarse subgrid with step $st_{wien}$.

\paragraph{Estimation.}
Given a reference patch $\ma p$, a group of similar patches selected as described in Section \ref{sec:patch_search}, but using patches extracted from the basic estimate for computing the patch distance. Two sets of patches are extracted: one from the noisy sequence and the other one from the basic estimate at the same locations. Once the sets of candidates, $\mathcal{P}(\ma p)$ from the noisy sequence and $\mathcal{P}(\hat{\ma p})$ for the basic sequence, have been computed, a Wiener filtering step is applied. The coefficients for the filtering are computed using $\mathcal{P}(\hat{\ma p})$ but it's $\mathcal{P}\left( \ma p \right)$ which is used for the estimation. Just like the first step, a 3D domain transform is first applied to the group of patches before the application of the Wiener filter. The computation is done following Equation \eqref{eq:wiener_filtering}. 
\begin{equation}
	\widehat{\mathcal{P}(\ma p)} = T_{3D}^{-1}(WF(T_{3D}(\mathcal{P}(\ma p))))
	\label{eq:wiener_filtering}
\end{equation}
where $WF$ on a frequency $\nu$ (with $\hat{\nu}$ corresponding to the same frequency in the basic estimation) is defined by
\begin{equation}
	WF(\nu) = \frac{\hat{\nu}^2}{\hat{\nu}^2+\sigma^2}  f\\
	\label{eq:WF}
\end{equation}
In practice the 3D transforms are chosen as a 2D transform applied directly on the patches of $\mathcal{P}(\ma p)$ and $\mathcal{P}(\hat{\ma p})$, typically a DCT, followed by an 1D transform along the third dimension of the group, typically a Haar transform.
The pseudocode of the estimation for the first step is presented in Algorithm \ref{alg:wt}.

\paragraph{Aggregation.}
Just as with the first step, the estimation gives multiple estimates per pixel. Therefore the different estimates are aggregated  using the same principle as in Section \ref{sec:first_step}, defined by Equation \eqref{eq:aggregation}. The only difference lies in the weights.
The weights $\omega$ used during the aggregation are computed in part during the estimation. The part $w_{wf}$ computed during the estimation is $w_{wf}(\ma p) = \frac{1}{\sigma^2} \left(\sum_{\rho}^{} \left(\frac{\hat{\ma p}_i(\rho)^2}{\hat{\ma p}_i(\rho)^2 + \sigma^2}\right)^2\right)^{-1}$ which is actually linked to the squared $\ell_2$ norm of the vector of coefficients used for the filtering. The rest of the weight come from a 2D Kaiser window $K(\ma p)$ of size $k_{wien}\times k_{wien}$ applied to avoid boundary effects from the patches.
Finally the pixel weight is defined by $\omega_{\ma p}(x) = w_{wf}(\ma p) K(\ma p)$. The pseudocode with the aggregation step can be found in Algorithm \ref{alg:vbm3d}.

\begin{algorithm}
	\caption{VBM3D algorithm}
	\label{alg:vbm3d}
	\setstretch{1.4}
	\DontPrintSemicolon
	\Input{A video $v$, the noise variance $\sigma^2$, the number of similar patches to compute $N$, a number of temporal frames $N_f$, the size of the search region in the reference frame $N_s$, the size of the search region in the other frame $N_{pr}$ the number of patches kept in each frame $N_b$, the size of the patch $k$, $d$, the distance threshold $\tau$, the thresholding parameter $\lambda_{3D}$, the domain transform $T_{3D}$, the coefficient fo the Kaiser window $\beta$, the step of the grid on which the patch are taken $st$}
	\Output{An final estimate denoised video $\hat{v}^{(2)}$}
	$K_1 \leftarrow $ Kaiser window of size $k_{hard}$ and coefficient $\beta_{hard}$\;
	$K_2 \leftarrow $ Kaiser window of size $k_{wien}$ and coefficient $\beta_{wien}$\;
	\tcp{Step 1}
	\For{each $\ma p$ on the grid of step $st_{hard}$}{
		\tcp{Search for similar patches in the noisy video}
		$L_{\ma p} \leftarrow \texttt{compute\_similar\_patches}(v, \ma p, N_{hard}, N_f, N_s, N_{pr}, N_b, k_{hard}, d_{hard}, \tau_{hard})$\;	
		\tcp{Filter the group of patches using a hard thresholding}
		$(\hat{L_{\ma p}}, \omega) \leftarrow \texttt{ht\_filtering}(L_{\ma p}, T_{3D, hard},  \sigma^2)$\;
		\For{$\ma q \in \hat{L_{\ma p}}$}{
			$a(\ma q) \leftarrow a(\ma q) + \omega K_1 \otimes \ma q$\;
			$w(\ma q) \leftarrow w(\ma q) + \omega K_1$\;
		}
	}
	\For{each pixel $x$}{
		$\hat{v}^{(1)}(x) \leftarrow a(x) / w(x)$\;
	}
	\tcp{Step 2}
	\For{each $\ma p$ on the grid of step $st_{wien}$}{
		\tcp{Search for similar patches in the basic estimate}
		$L'_{\ma p} \leftarrow \texttt{compute\_similar\_patches}(\hat{v}^{(1)}, \ma p, N_{wien}, N_f, N_s, N_{pr}, N_b, k_{wien}, d_{wien}, \tau_{wien})$\;	
		$L_{\ma p} \leftarrow $ patches from $v$ at the same position than the one from $L'_{\ma p}$\;
		\tcp{Filter the group of patches using a Wiener filtering}
		$(\hat{L_{\ma p}}, \omega) \leftarrow \texttt{wiener\_filtering}(L_{\ma p}, L'_{\ma p}, T_{3D, wien}, \sigma^2)$\;
		\For{$\ma q \in \hat{L_{\ma p}}$}{
			$a(\ma q) \leftarrow a(\ma q) + \omega K_2 \otimes \ma q$\;
			$w(\ma q) \leftarrow w(\ma q) + \omega K_2$\;
		}
	}
	\For{each pixel $x$}{
		$\hat{v}^{(2)}(x) \leftarrow a(x) / w(x)$\;
	}
	\Return $\hat{v}^{(2)}$\;

\end{algorithm}

\subsection{Reproducing the original VBM3D}

We now compare the results obtained with our implementation to those obtained from the binaries released by the authors of the original VBM3D \cite{Dabov2007v}\footnote{Available at \url{http://www.cs.tut.fi/~foi/GCF-BM3D/}.}.
Throughout this work we will use a test set of 
seven grayscale test sequences obtained from the \textit{Derf's Test Media collection}\footnote{\url{https://media.xiph.org/video/derf/}.}. The original sequences are of higher resolution and in RGB. They have been downscaled and converted to grayscale.

Table \ref{tab:ori_vs_ours} compares the PSNR obtained by our implementation with the original binaries. Our results below are those of the original implementation. The average gap starts at 0.27dB for $\sigma=10$ and reduces to 0.18dB for $\sigma = 40$. \emph{Station} and \emph{sunflower} are the sequences where the gap is larger, reaching 0.68dB for the later at $\sigma= 10$.

\begin{figure}[ht!]
    \centering
    \includegraphics[width=0.33\linewidth,trim={10cm 4cm 10cm 4cm}, clip]{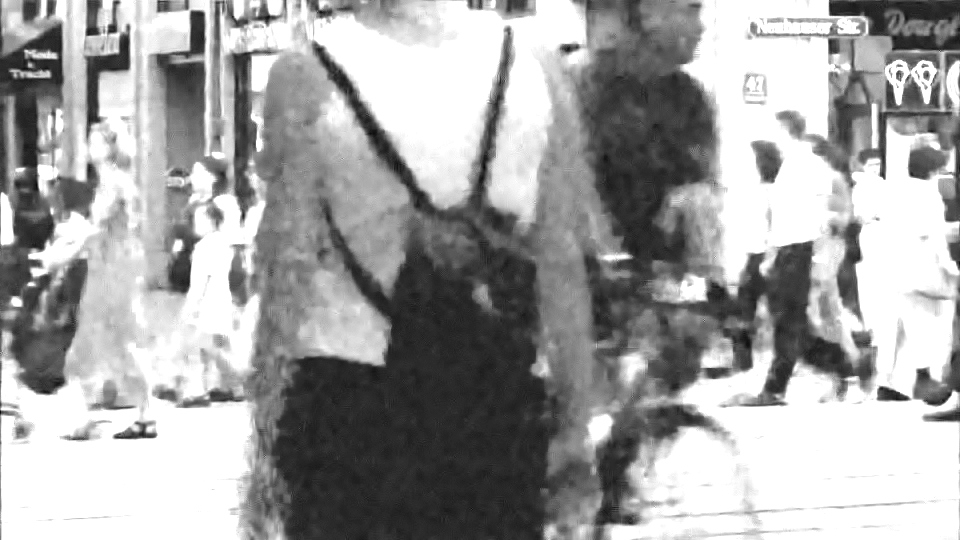}%
    \includegraphics[width=0.33\linewidth,trim={10cm 4cm 10cm 4cm}, clip]{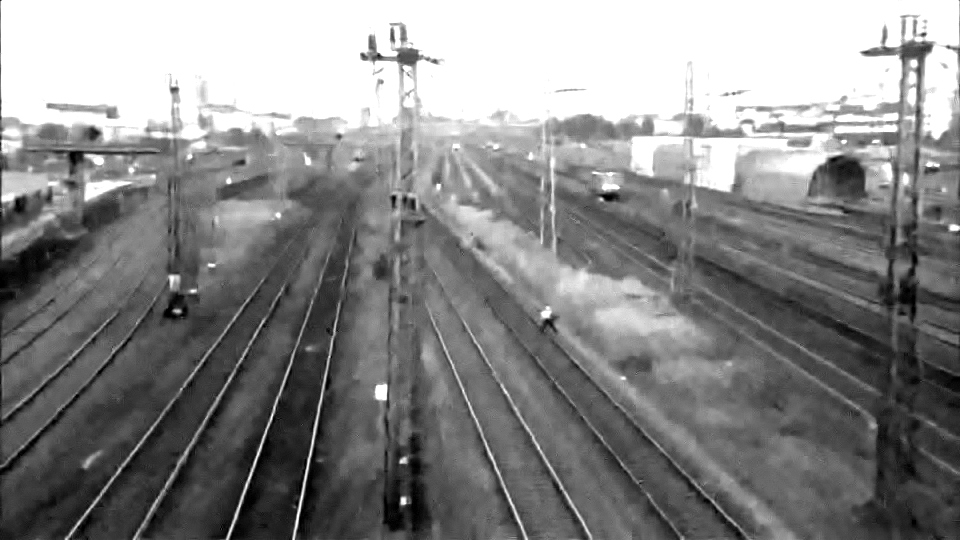}%
    \includegraphics[width=0.33\linewidth,trim={ 5cm 3cm 15cm 5cm}, clip]{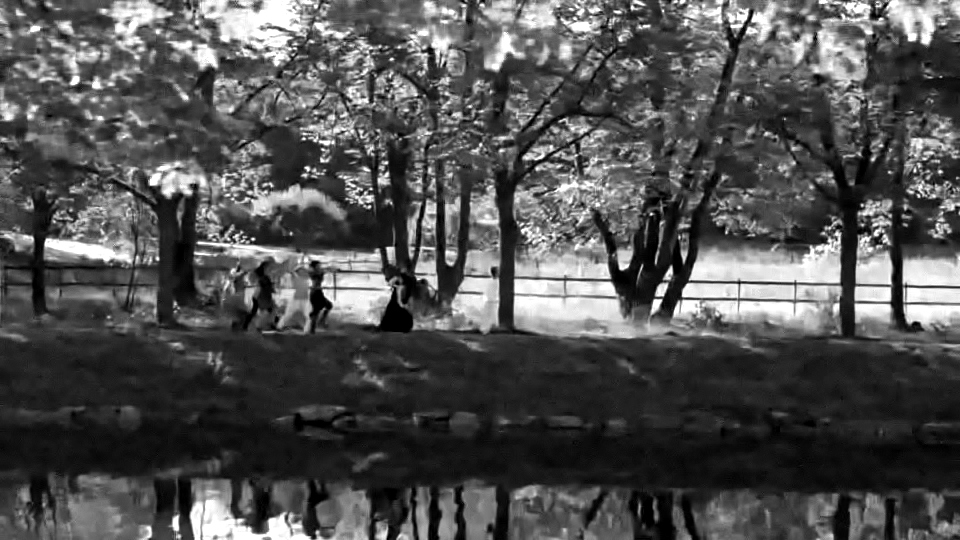}%

    \includegraphics[width=0.33\linewidth,trim={10cm 4cm 10cm 4cm}, clip]{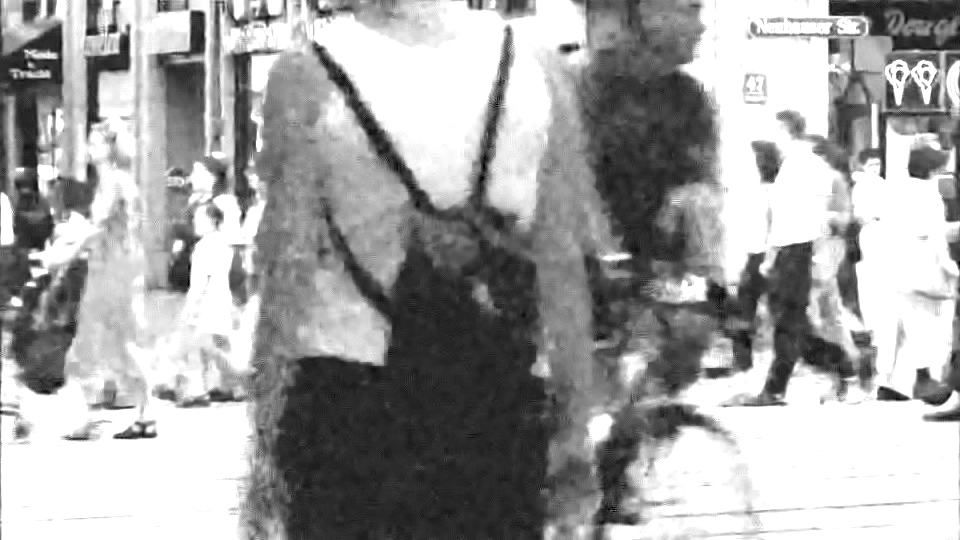}%
    \includegraphics[width=0.33\linewidth,trim={10cm 4cm 10cm 4cm}, clip]{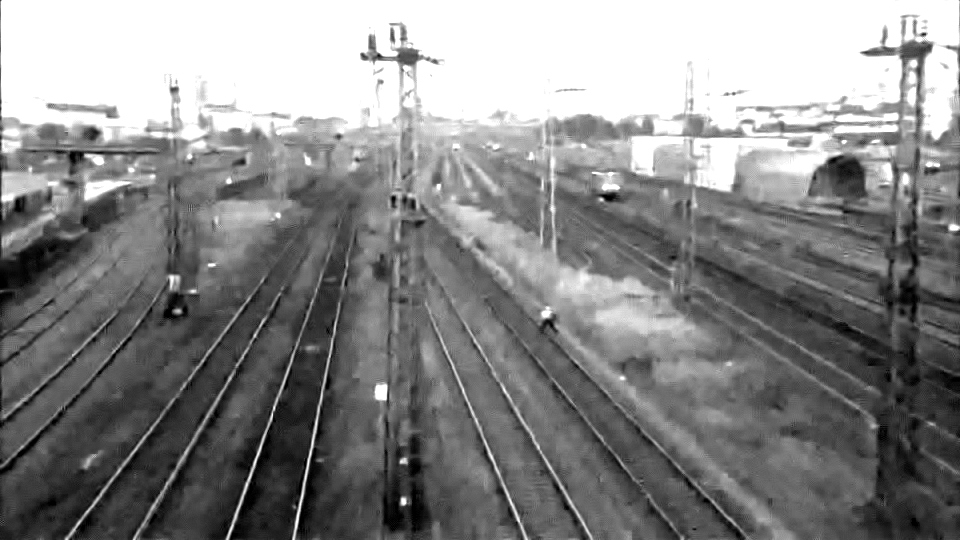}%
    \includegraphics[width=0.33\linewidth,trim={ 5cm 3cm 15cm 5cm}, clip]{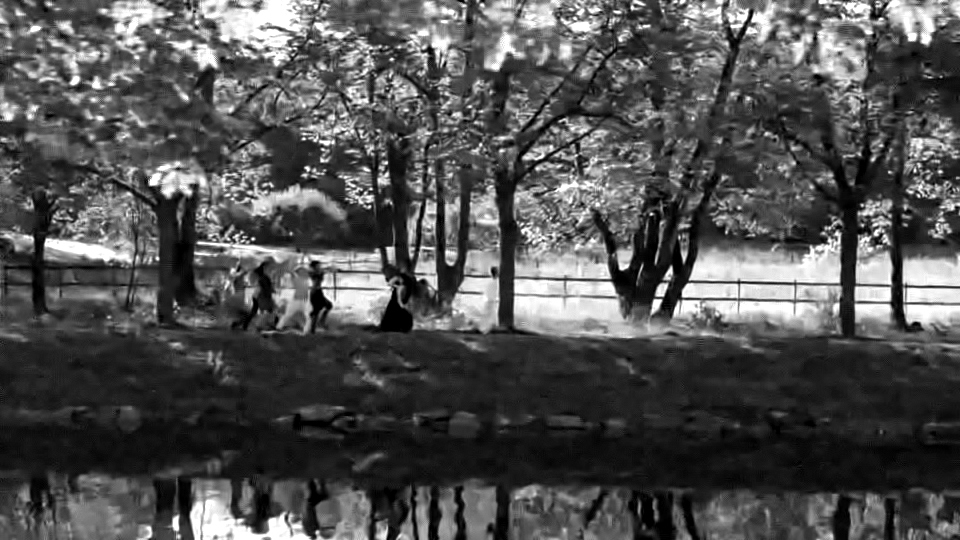}%

	 \caption{Top: results of  VBM3D \cite{Dabov2007v}, with the original authors' implementation. Bottom: result 	obtained with our implementation. The noise level is $\sigma=40$. The
		contrast has been linearly scaled for better visualization.
		}
	\label{fig:ori_vs_ours}
\end{figure}

A visual comparison of both results is shown in Figure \ref{fig:ori_vs_ours}, for noise $\sigma = 40$. Both results are visually very similar, with the same type of artifacts. A closer inspection reveals that the result of the original VBM3D is slightly smoother, resulting in less noticeable DCT artifacts and on some textures with decreased contrast (see for instance the grass in the rightmost figure).  

\begin{table*}
	\begin{center}
		{\small
		\renewcommand{\tabcolsep}{1.6mm}
		\renewcommand{\arraystretch}{1.0}
\resizebox{\linewidth}{!}{
		\begin{tabular}{@{} c l c c c c c c c @{\hskip 1cm} c @{}}
		\toprule
\rule{0pt}{10pt}$\sigma$ 
& Method           &    Crowd   &     Park   & Pedestrians & Station   &  Sunflower &  Touchdown &   Tractor  & Average    \\
\midrule
\multirow{1}{*}{$10$}                                                                  
& VBM3D (original) &    35.65   &    34.75   &    40.83   &    38.93   &    40.49   &    39.04   &    37.01   &    38.10   \\
& VBM3D (ours)     &    35.52   &    34.59   &    40.65   &    38.38   &    39.81   &    39.01   &    36.82   &    37.83   \\
\midrule
\multirow{1}{*}{$20$}                                 
& VBM3D (original) &    32.25   &    31.25   &    36.94   &    35.45   &    36.46   &    36.08   &    33.07   &    34.50   \\
& VBM3D (ours)     &    32.06   &    31.12   &    36.81   &    35.10   &    35.95   &    36.05   &    32.97   &    34.30   \\
\midrule
\multirow{1}{*}{$40$}   
& VBM3D (original) &    28.65   &    27.68   &    32.81   &    32.02   &    32.65   &    33.52   &    29.41   &    30.96   \\
& VBM3D (ours)     &    28.39   &    27.64   &    32.62   &    31.80   &    32.31   &    33.35   &    29.38   &    30.78   \\
\bottomrule
		\end{tabular}}
}
	\end{center}
	\caption{Comparison of the denoising quality with the binary program provided by Dabov \etal with \cite{Dabov2007}. Results were both computed using the normal profile 'np' parameters.}
	\label{tab:ori_vs_ours}
\end{table*}

\section{Extensions}
\label{sec:extensions}

\subsection{Using optical flow to guide the search}
\label{sec:guided}

Many video denoising methods take advantage of the optical flow to estimate motion in the video. Typically, optical flow is used by video denoising methods that require aggregating information along motion trajectories \cite{Brailean1995,Liu2010,buades2016patch,ehret2018non,bwd-nlkalman-19}. These methods require a motion estimate as accurate as possible. Most patch-based methods, on the other hand, do not require such an accurate motion estimate, as they are based on finding similar patches in a 3D search region \cite{Buades2005v,Dabov2007v,Maggioni2012}. These methods either do not use any motion estimate at all, or use a very rough one (e.g. block matching) to guide the search region. However, it has been observed that using optical flow to shape the search region can still be beneficial for patch based methods \cite{Arias2018,arias-18-patch-models}, as it allows to find better matches.

For a pair of two images A and B, the optical flow aims at finding a vector field $o(x,y) = (\delta_x, \delta_y)$ such that any point $(x,y)$ of the image domain solves 
\begin{equation}
    A(x+\delta_x, y+\delta_y) = B(x,y).
\end{equation} 
The displacement vector can be sub-pixel, in which case the image $A$ needs to be interpolated.
We decided to use the TVL1 optical flow method \cite{zach2007duality}, in particular the implementation provided in \cite{ipol.2013.26}. We compute the optical flow in a downscaled version of the video by a factor of $4$, and scale it back to the original resolution. This reduces the running time and the impact of the noise while still having a reasonable precision, as shown in \cite{ehret2018non}.

We add the optical flow to VBM3D as a guide, the same way it is done for VNLB \cite{Arias2018}. 
The spatio-temporal search region is defined as two sequences of $N_f$ square windows
of size $N_{pr}\times N_{pr}$ (plus the $N_s \times N_s$ window corresponding to the reference frame), whose centers follow the motion trajectory of the
reference patch. The trajectory is estimated using the forward and backward optical flow. We use the center corresponding to the position of the reference patch propagated in previous, respectively following, frames using the backward, respectively forward, optical flow.
The forward half of the trajectory $\varphi_{x,y,t}$ passing through $(x,y,t)$ is computed by integrating the forward optical flow $o^f$ (also indexed by time on top of the spatial position) as follows:
\begin{equation}
\varphi_{x,y,t}(h) = o^f([\varphi_{x,y,t}(h-1)],h-1) + \varphi_{x,y,t}(h-1),\quad h = t+1,\dots,t + N_f,
\end{equation}
where $[\,\cdot\,]$ and $\lfloor\cdot\rfloor$ denote the round and floor
operators. The backward half of the trajectory is defined analogously using
the backward optical flow $o^b.$

This also means that only one center needs to be tracked, in contrast to the $N_b$ required by a regular VBM3D search. Parameters are kept the same as in the original algorithm.
The guided version of the search is summarized in Algorithm \ref{alg:vbm3d_guided_search}.

We also tested setting the parameter $d$ to zero for both steps since one can assume that it would be redundant with the regularization of the patch search offered by the optical flow. PSNR results are shown in Table~\ref{tab:psnr-guided}. Using the optical flow as a guide greatly improves the quality of the denoising. The non-zero $d$ yields better results both with and without the optical flow guide, thus we decided to keep it in our final version.

Figure \ref{fig:visual} shows results obtained with the different extensions that will be described in this section. The first column corresponds to VBM3D with the default parameters, and the 3rd column to the one using an optical flow guided search region (denoted ``VBM3D OF'').
Guiding the patch search allows to recover more details. This is because the tracking of patches is more robust to motion than the block-matching suggested for the original VBM3D and therefore provides better matches. Since the optical flow is used only as guide for the center of the search region, computing the optical flow from the noisy data is not a problem as it is not required to be very precise. 

\begin{algorithm}
	\caption{\texttt{compute\_similar\_patches}: VBM3D patch search guided by the optical flow}
	\label{alg:vbm3d_guided_search}
	\setstretch{1.4}
	\DontPrintSemicolon
	\Input{Noisy video $v = (v_0, \dots, v_f)$, a reference patch $\ma p$, the number of patches to return $N$, a number of temporal frames $N_f$, a search region for the reference frame $N_s$, a search region for the other frames $N_{pr}$, the maximum number of patches to compute with each local search $N_b$, the size of the patch $k$, a correcting factor $d$, a maximum distance threshold $\tau$, the forward and backward optical flows $o_f$ and $o_b$}
	\Output{$L$ the list of the $N$ patches closest to $\ma p$ and their distance}
	$t \leftarrow$ frame at which $\ma p$ is located\;
	$L_{t} \leftarrow \texttt{local\_search}(\ma p, \ma p, N_s, k, N_b, d, u_t)$\;
	\tcp{Search in the $N_f$ following frames}
	$\ma q \leftarrow \ma p$\;
	\For{$t_f = (t+1)$ to $\min(t+N_f, f-1)$}{
	    $\ma q \leftarrow o_f(\ma q, t_f-1)$ \Comment*{Follow the center using the forward optical flow}
		 $L_{t_f} \leftarrow \texttt{local\_search}(\ma q, \ma p, N_{pr}, k, N_b, d, u_{t_f})$\; 
	}
	\tcp{Search in the $N_f$ previous frames}
	$\ma q \leftarrow \ma p$\;
	\For{$t_p = t-1$ to $\max(t-N_f, 0)$}{
	    $\ma q \leftarrow o_b(\ma q, t_p+1)$ \Comment*{Follow the center using the forward optical flow}
		 $L_{t_p} \leftarrow \texttt{local\_search}(\ma q, \ma p, N_{pr}, k, N_b, d, u_{t_p})$\; 
	}
	$L \leftarrow \bigcup_{i \in \llbracket t-N_f; t+N_f \rrbracket} L_i$\;
	$L \leftarrow $ elements from $L$ with distance smaller than $\tau$\;
	\tcp{the 3D transform requires a number of patch which is a power of $2$}
	$N_l \leftarrow $closest power of $2$ small or equal than $min(size of L, N)$\;
	\If{$L$ has more than $N_l$ elements}{
		$L \leftarrow N_l$ best candidate from $L$\;
	}
	\Return $L$\;
\end{algorithm}

\begin{table*}
	\begin{center}
		{\small
		\renewcommand{\tabcolsep}{1.6mm}
		\renewcommand{\arraystretch}{1.0}
\resizebox{\linewidth}{!}{
		\begin{tabular}{@{} c l c c c c c c c @{\hskip 1cm} c @{}}
		\toprule
\rule{0pt}{10pt}$\sigma$ 
& Method               &     Crowd &     Park   & Pedestrians &   station &   Sunflower & Touchdown &    Tractor &    Average \\
\midrule
\multirow{1}{*}{$10$}                                                                  
& VBM3D (without)     &      35.52 &      34.59 &      40.65 &      38.38 &      39.81 &      39.01 &      36.82 &      37.83 \\
& VBM3D (without,d=0) &      35.72 &      35.00 &      40.13 &      39.34 &      40.18 &      39.01 &      36.94 &      38.05 \\
& VBM3D (with)        &      35.61 &      34.94 &\Best{41.04}&      39.79 &      41.75 &\Best{39.89}&      38.43 &      38.78 \\
& VBM3D (with,d=0)    &\Best{35.78}&\Best{35.19}&      40.61 &\Best{40.27}&\Best{41.85}&      39.75 &\Best{38.51}&\Best{38.85}\\
\midrule
\multirow{1}{*}{$20$}                                 
& VBM3D (without)     &      32.06 &      31.12 &      36.81 &      35.10 &      35.95 &      36.05 &      32.97 &      34.30 \\
& VBM3D (without,d=0) &      32.00 &      31.24 &      36.13 &      35.38 &      36.09 &      35.83 &      33.03 &      34.24 \\
& VBM3D (with)        &\Best{32.17}&      31.44 &\Best{37.32}&      36.26 &\Best{38.02}&\Best{36.91}&\Best{34.58}&\Best{35.24}\\
& VBM3D (with,d=0)    &      32.10 &\Best{31.49}&      36.72 &\Best{36.32}&      37.98 &      36.61 &\Best{34.58}&      35.11 \\
\midrule
\multirow{1}{*}{$40$}   
& VBM3D (without)     &      28.39 &      27.64 &      32.62 &      31.80 &      32.31 &      33.35 &      29.38 &      30.78 \\
& VBM3D (without,d=0) &      28.29 &      27.65 &      32.24 &      31.79 &      32.33 &      33.11 &      29.40 &      30.69 \\
& VBM3D (with)        &\Best{28.55}&\Best{27.99}&\Best{33.29}&\Best{32.80}&\Best{34.46}&\Best{34.00}&\Best{30.79}&\Best{31.70}\\
& VBM3D (with,d=0)    &      28.45 &      27.98 &      32.94 &      32.75 &      34.40 &      33.79 &      30.76 &      31.58 \\
\bottomrule
		\end{tabular}}
		}
	\end{center}
   \caption{Comparison of the denoising quality with and without guiding with an optical flow. Guiding the search leads to much better results. It is also better in general to keep the additional regularization given by $d$.}
	\label{tab:psnr-guided}
\end{table*}

\subsection{Spatio-temporal patches}
\label{sec:spatiotemp}

The patch similarity is determined based on the (squared) Euclidean distance 
between patches. Due to the noise, the Euclidean distance follows a non-central
$\chi^2$ distribution, with variance
\[\textnormal{var}\left\{\frac1{m}\|\ma q_1 - \ma q_2\|^2\right\} = \frac{8\sigma^2}{m}\left(\sigma^2 + \frac1{m}\|\ma p_1 - \ma p_2\|^2\right),\]
where $\ma q_1, \ma q_2$ are the noisy versions of the patches $\ma p_1, \ma p_2$, 
and $m$ denotes the number of pixels in the patch.
The noise in the distance
can be reduced by considering larger patches.
However, increasing the size of the patch also increases the distances between
patches and reduces the likelihood of finding similar ones.
The additional temporal dimension in a spatio-temporal patch allows to increase
the number of pixels in the patch, without increasing its spatial size.
Due to the high redundancy of the video in the temporal dimension, increasing the 
temporal size of the patch causes a much lower increase in the patch distances.

When the motion is known or can be estimated, then it is natural to consider
motion compensated spatio-temporal patches (see for instance \cite{Maggioni2012,buades2016patch}). 
Alternatively, rectangular spatio-temporal patches with no motion compensation have been also used \cite{Protter2009,Arias2018}. 
For more complex types of motion, using rectangular spatio-temporal patches will result in
a larger variability in the set of nearest neighbors of a given patch, due to the fact that
both the spatial texture and the motion pattern may vary. At least in principle, better 
results should be obtained using motion compensation. 
However, in practice, for
higher levels of noise the bad quality of the estimated motion can undermine the
final result.

The principle of BM3D has been applied to 3D patches with and without motion compensation.
VBM4D, introduced in \cite{Maggioni2012}, uses motion compensated spatio-temporal patches for video denoising (the ``V'' stands for video). The motion is estimated using block matching.
BM4D uses cubic patches without motion compensation (of size $4\times 4\times 4$ or $5\times 5\times 5$), aiming at filtering volumetric images \cite{Maggioni2013}. In \cite{Maggioni2013} the authors compare the performance of both VBM4D and BM4D on video denoising concluding that VBM4D was the best video filtering strategy.

However, in \cite{Arias2018,arias-18-patch-models} it is shown that rectangular spatio-temporal patches with a temporal size of only two frames improve the denoising quality and still provide higher temporal consistency than a 2D patch. 
Based on those results, in this work we evaluate rectangular spatio-temporal patches of size $8\times 8\times 2$ in the first step and $7\times 7\times 2$ in the second (i.e we keep the spatial patch size).

In Table \ref{tab:of-st-psnr-classic-gray} we compare the quantitative results obtained by using spatial and spatio-temporal patches (denoted by ``ST'' in the table). We also consider the effect of the optical-flow-guided patch search, indicated as ``OF''.

\begin{landscape}
\begin{figure}
    \centering
    \includegraphics[width=0.166\linewidth,trim={4cm 7cm 16cm 0cm}, clip]{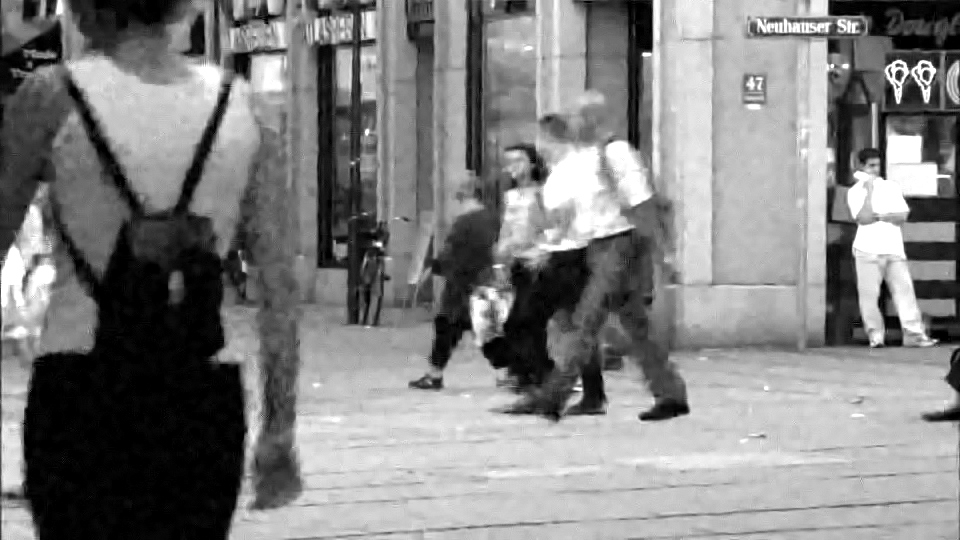}%
    \includegraphics[width=0.166\linewidth,trim={4cm 7cm 16cm 0cm}, clip]{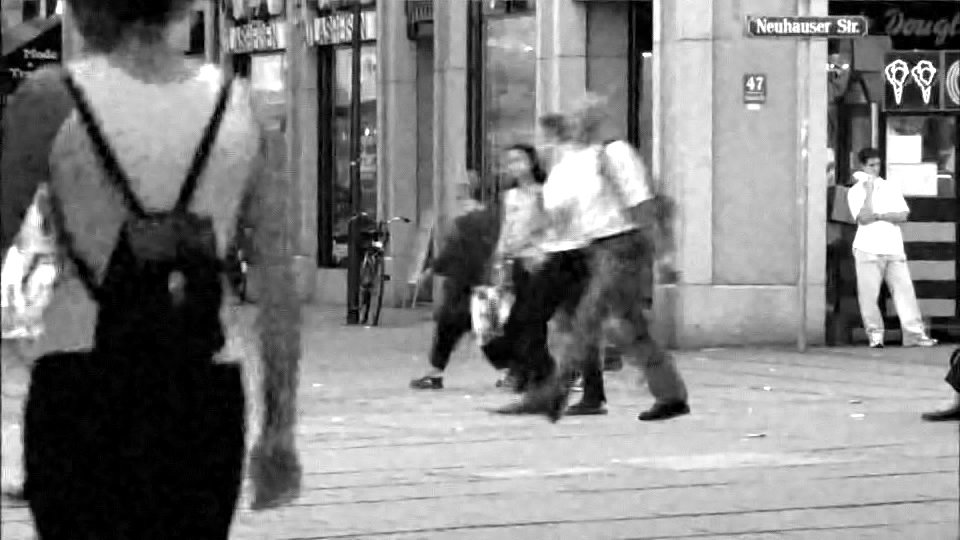}%
    \includegraphics[width=0.166\linewidth,trim={4cm 7cm 16cm 0cm}, clip]{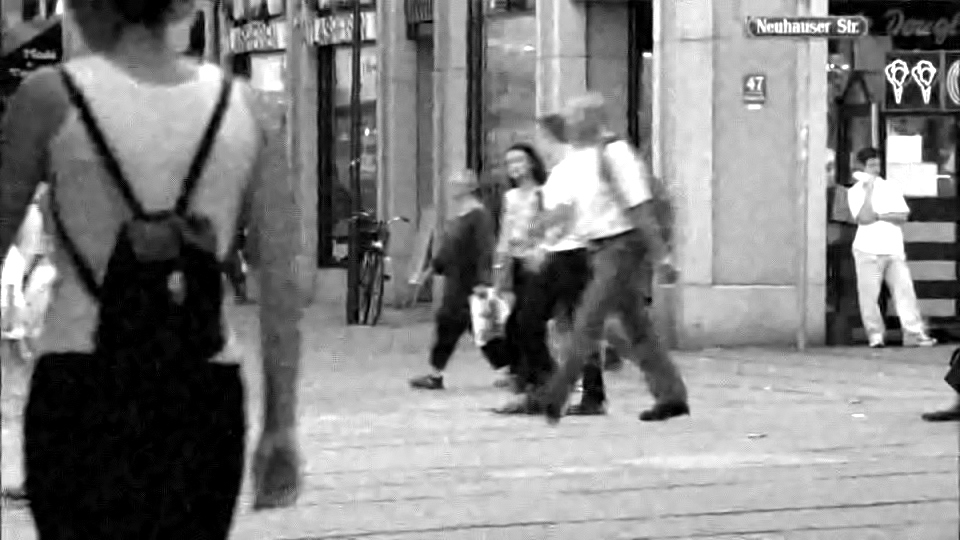}%
    \includegraphics[width=0.166\linewidth,trim={4cm 7cm 16cm 0cm}, clip]{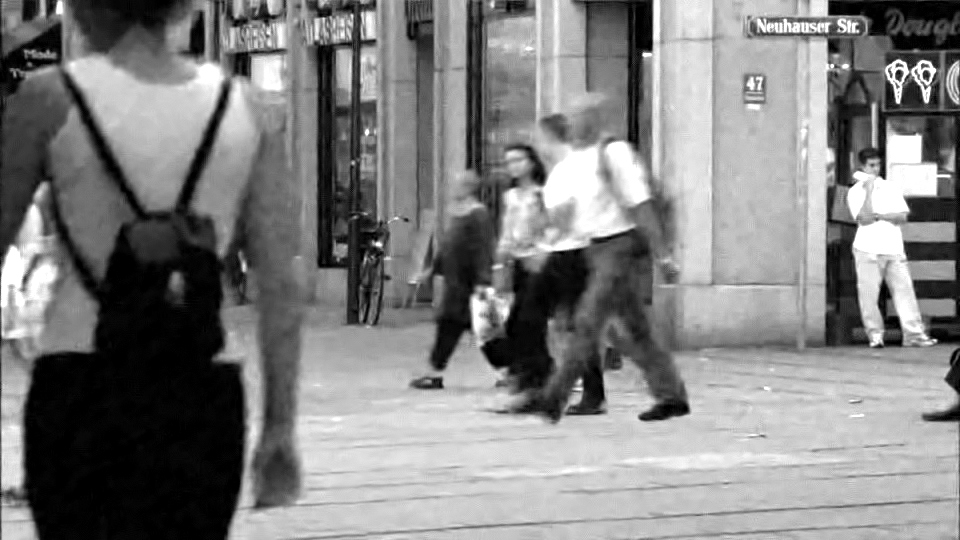}%
    \includegraphics[width=0.166\linewidth,trim={4cm 7cm 16cm 0cm}, clip]{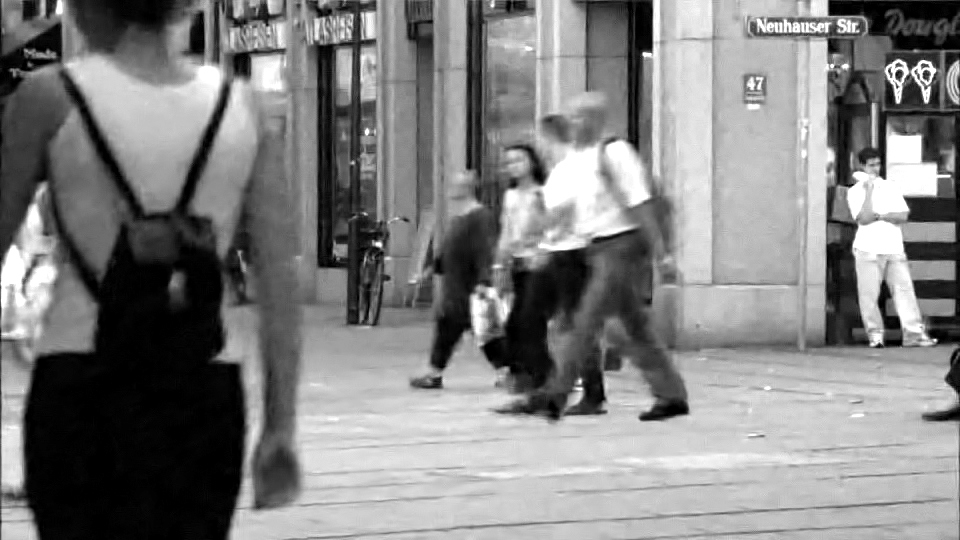}%
    \includegraphics[width=0.166\linewidth,trim={4cm 7cm 16cm 0cm}, clip]{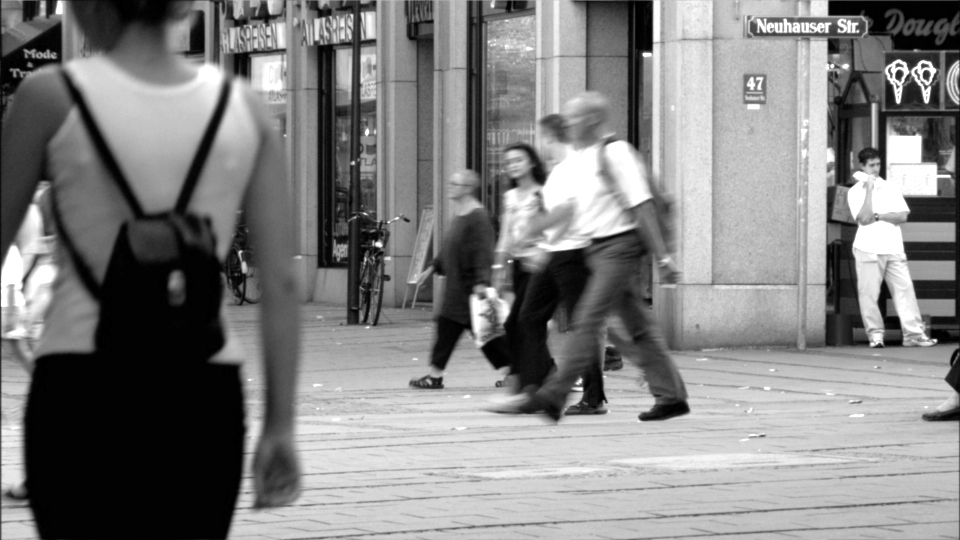}

    \includegraphics[width=0.166\linewidth,trim={16cm 0cm 0cm 4cm}, clip]{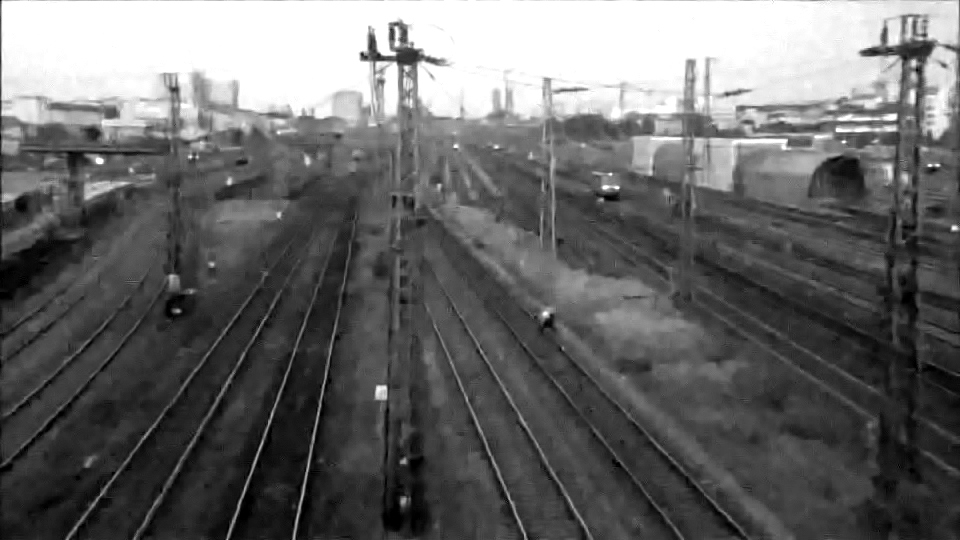}%
    \includegraphics[width=0.166\linewidth,trim={16cm 0cm 0cm 4cm}, clip]{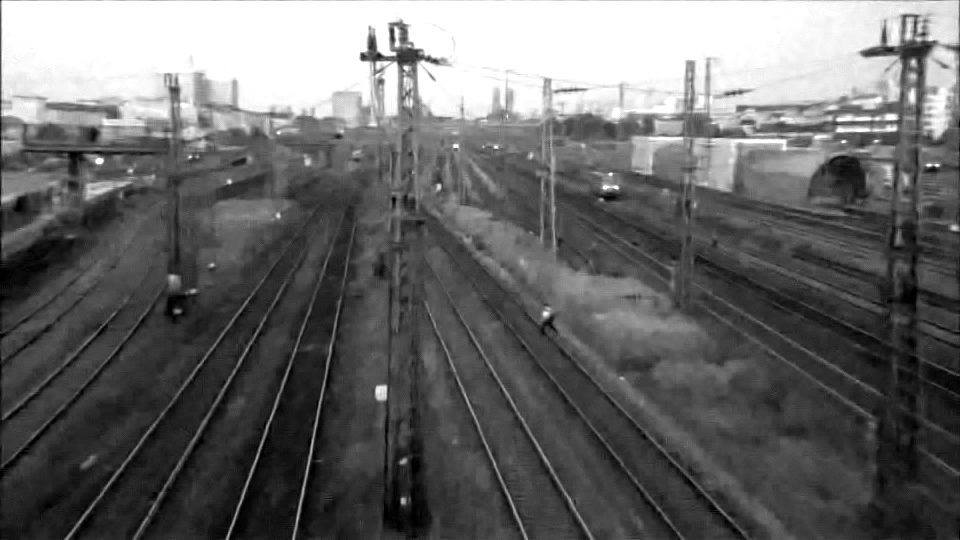}%
    \includegraphics[width=0.166\linewidth,trim={16cm 0cm 0cm 4cm}, clip]{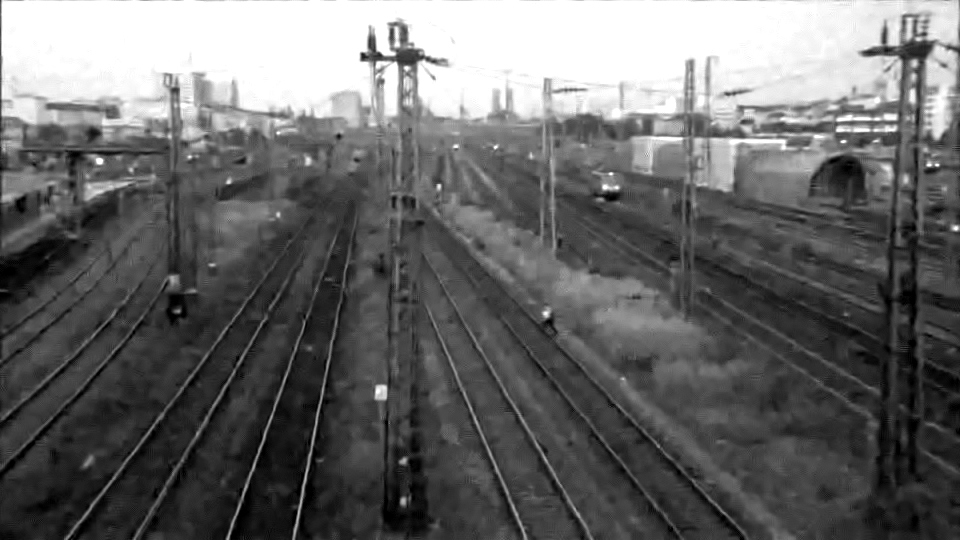}%
    \includegraphics[width=0.166\linewidth,trim={16cm 0cm 0cm 4cm}, clip]{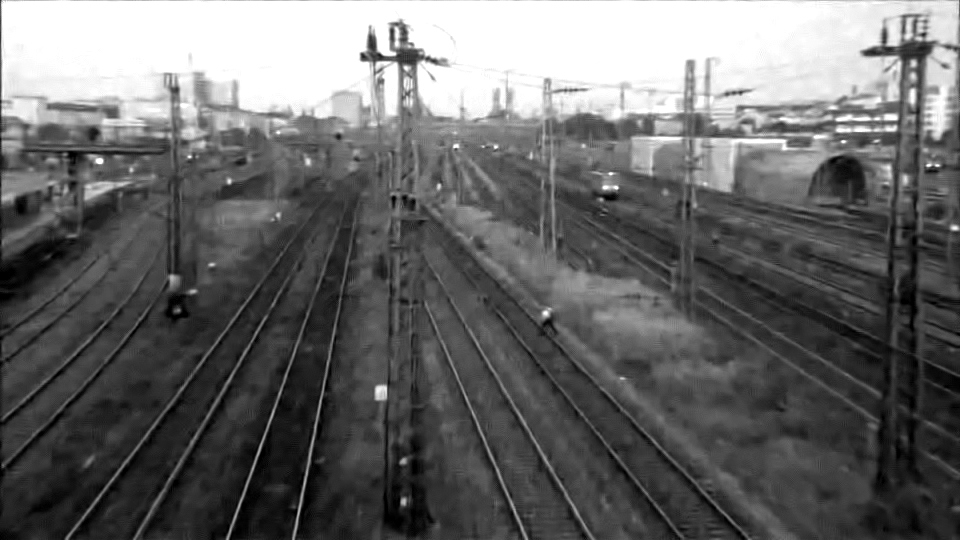}%
    \includegraphics[width=0.166\linewidth,trim={16cm 0cm 0cm 4cm}, clip]{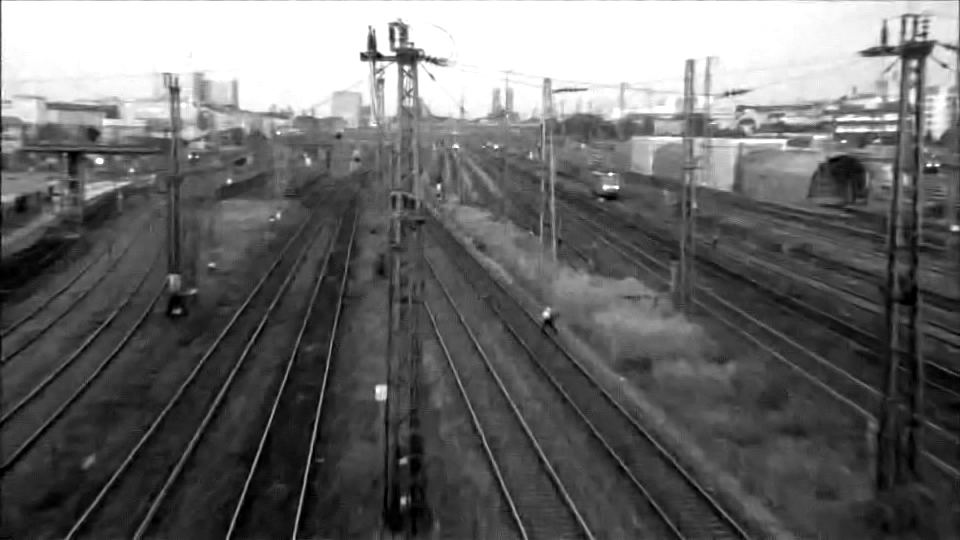}%
    \includegraphics[width=0.166\linewidth,trim={16cm 0cm 0cm 4cm}, clip]{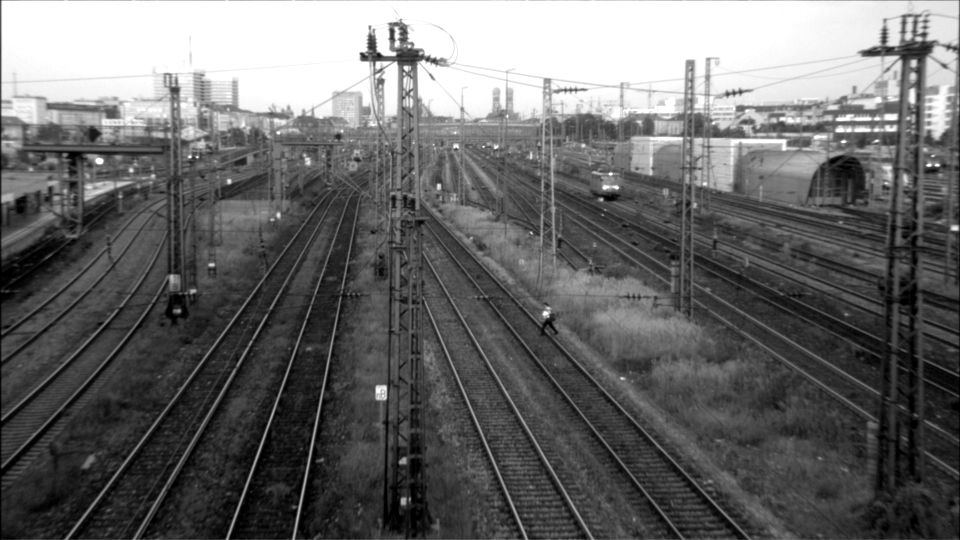}%

    \includegraphics[width=0.166\linewidth,trim={1cm 1cm 6cm 0cm}, clip]{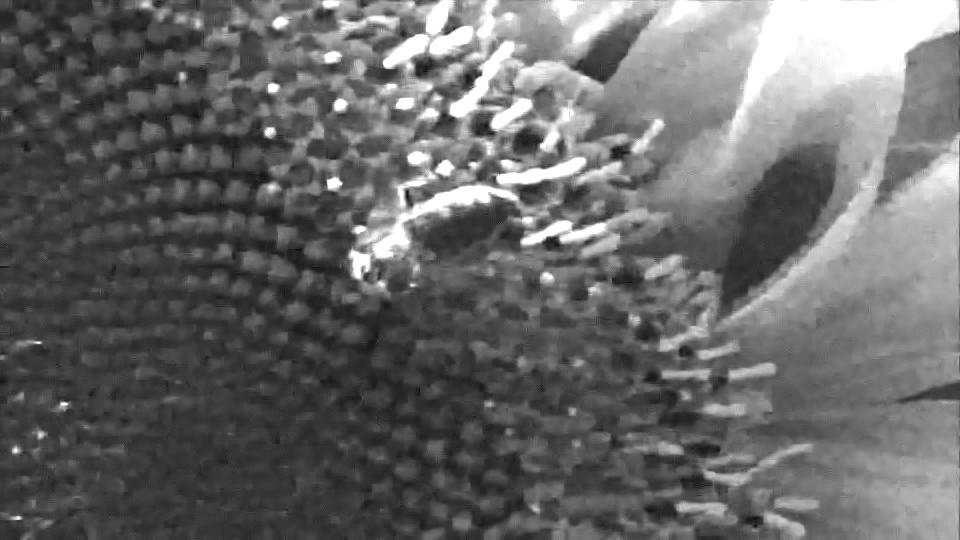}%
    \includegraphics[width=0.166\linewidth,trim={1cm 1cm 6cm 0cm}, clip]{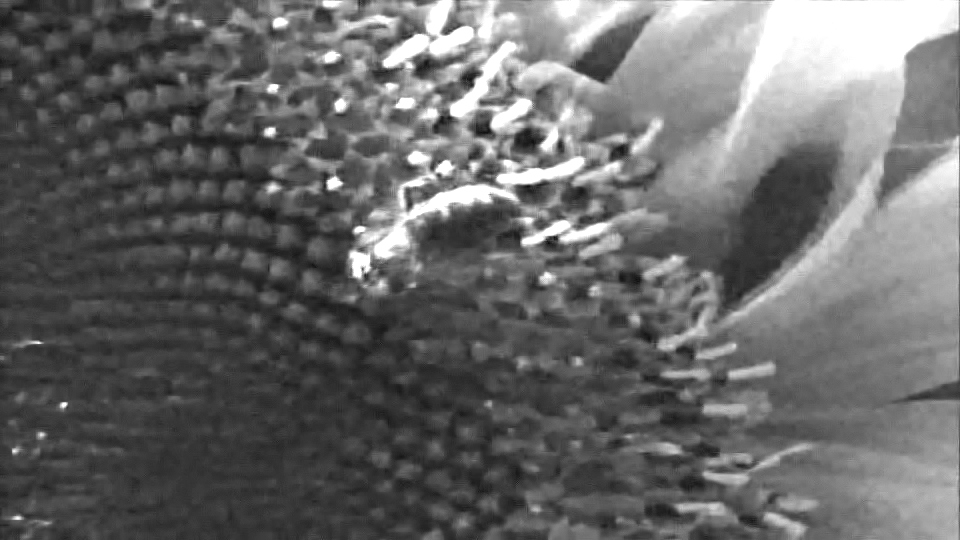}%
    \includegraphics[width=0.166\linewidth,trim={1cm 1cm 6cm 0cm}, clip]{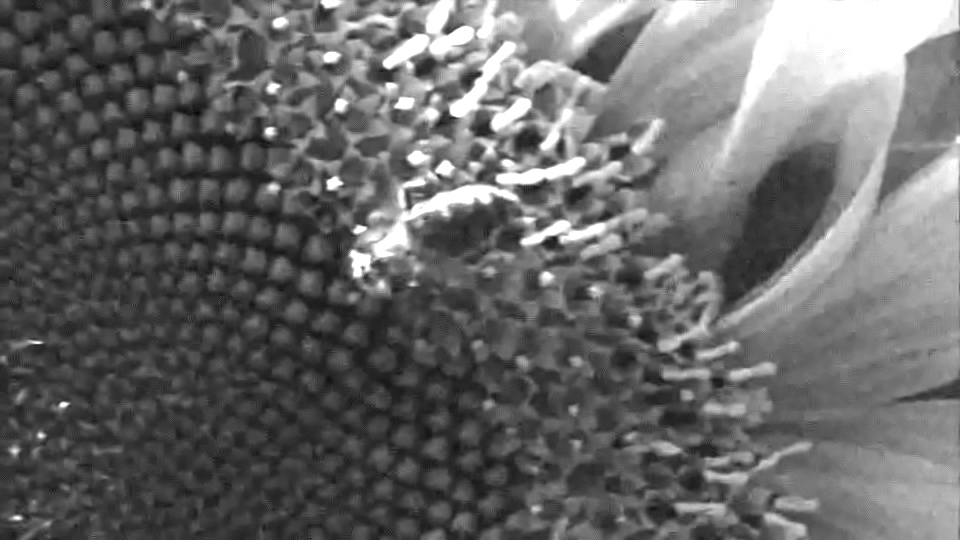}%
    \includegraphics[width=0.166\linewidth,trim={1cm 1cm 6cm 0cm}, clip]{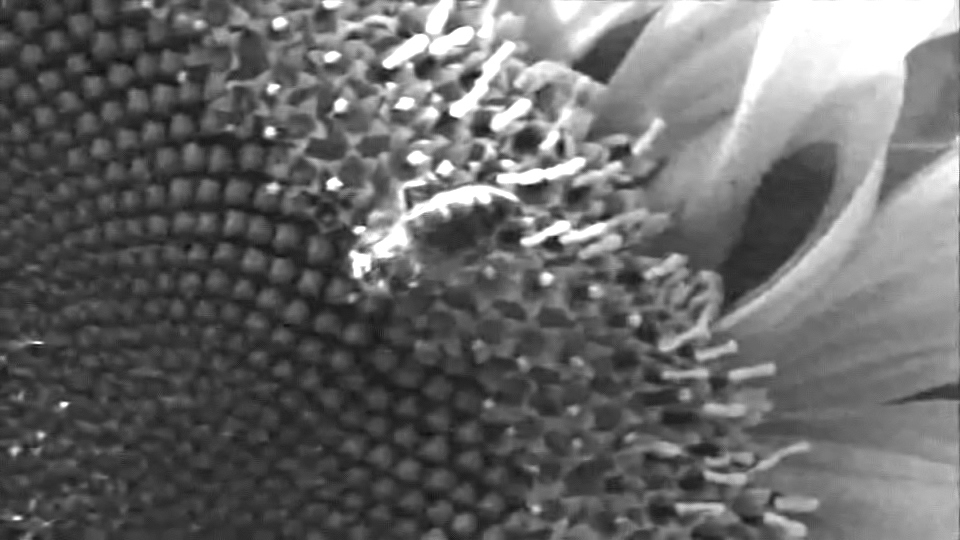}%
    \includegraphics[width=0.166\linewidth,trim={1cm 1cm 6cm 0cm}, clip]{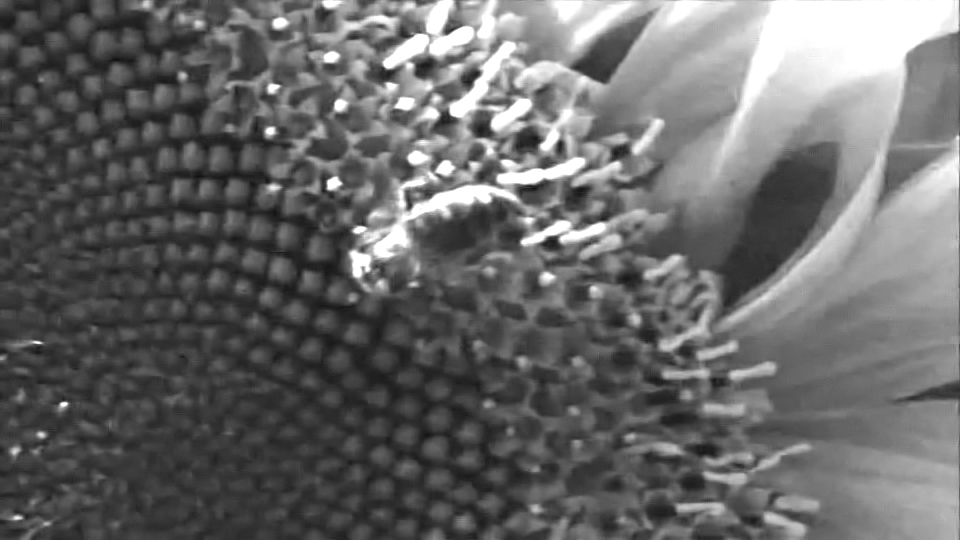}%
    \includegraphics[width=0.166\linewidth,trim={1cm 1cm 6cm 0cm}, clip]{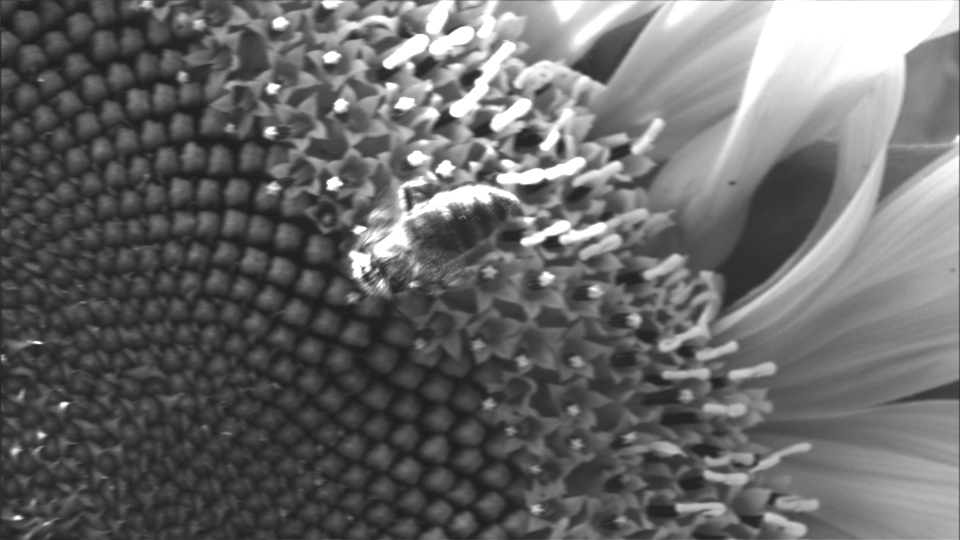}%

    \includegraphics[width=0.166\linewidth,trim={0cm 0cm 16cm 4cm}, clip]{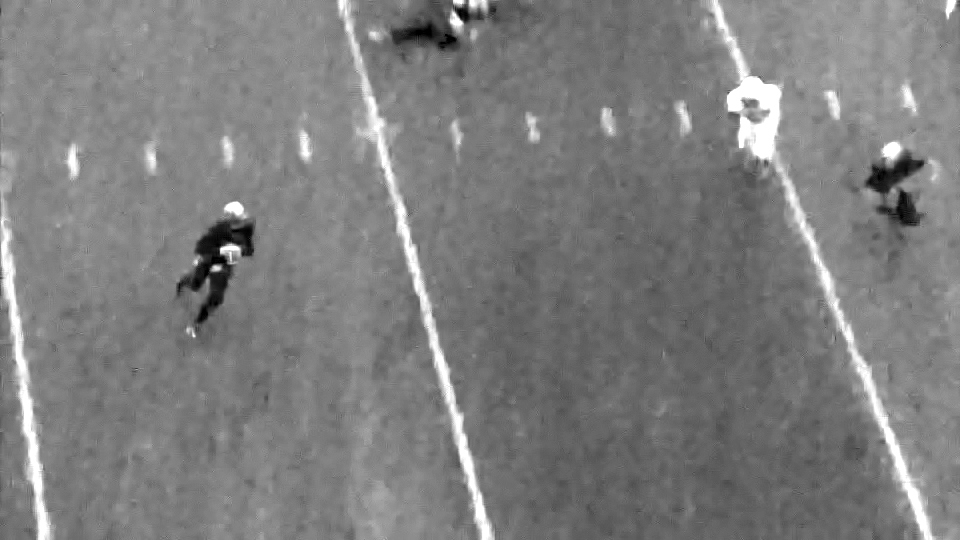}%
    \includegraphics[width=0.166\linewidth,trim={0cm 0cm 16cm 4cm}, clip]{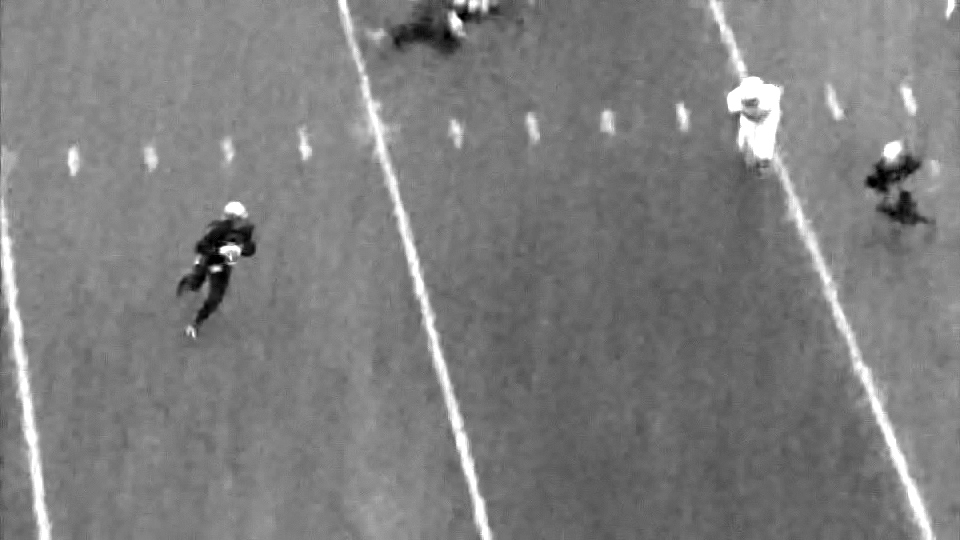}%
    \includegraphics[width=0.166\linewidth,trim={0cm 0cm 16cm 4cm}, clip]{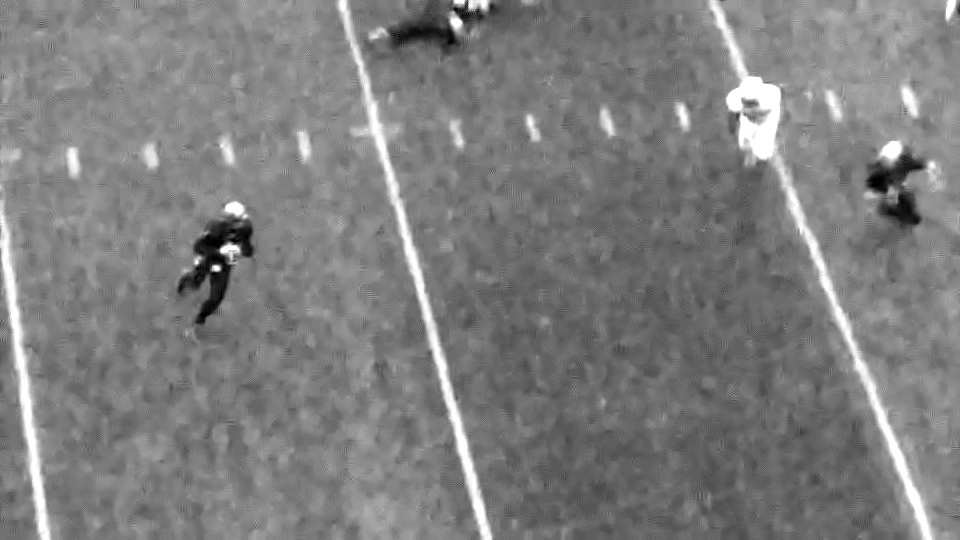}%
    \includegraphics[width=0.166\linewidth,trim={0cm 0cm 16cm 4cm}, clip]{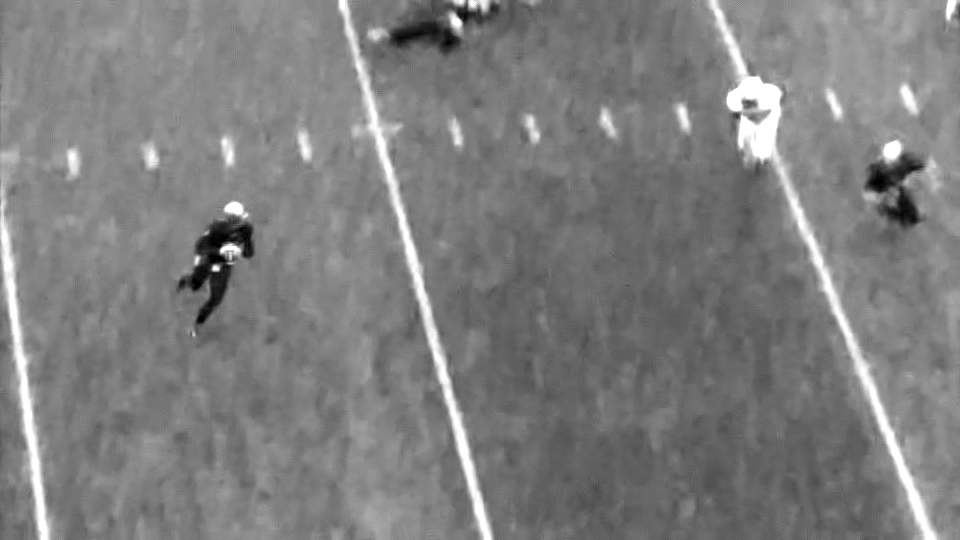}%
    \includegraphics[width=0.166\linewidth,trim={0cm 0cm 16cm 4cm}, clip]{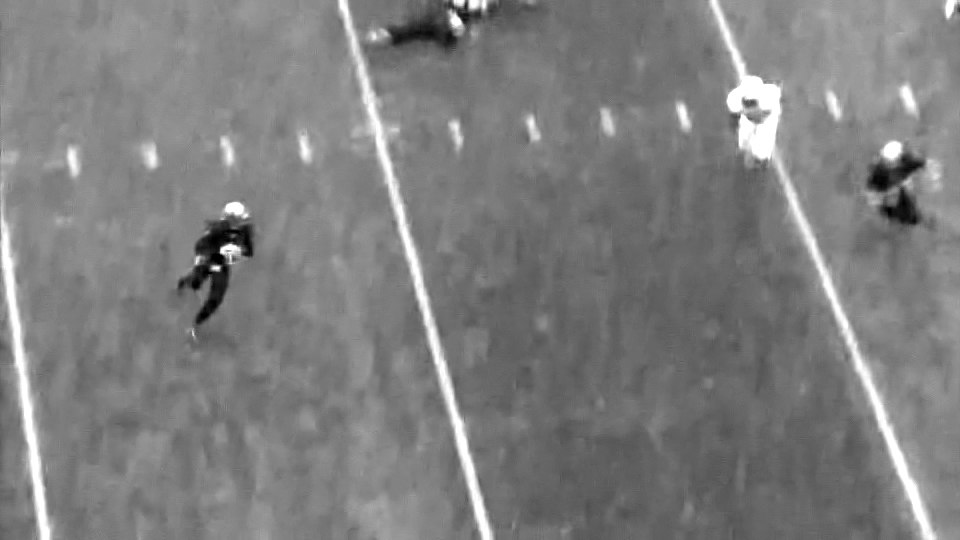}%
    \includegraphics[width=0.166\linewidth,trim={0cm 0cm 16cm 4cm}, clip]{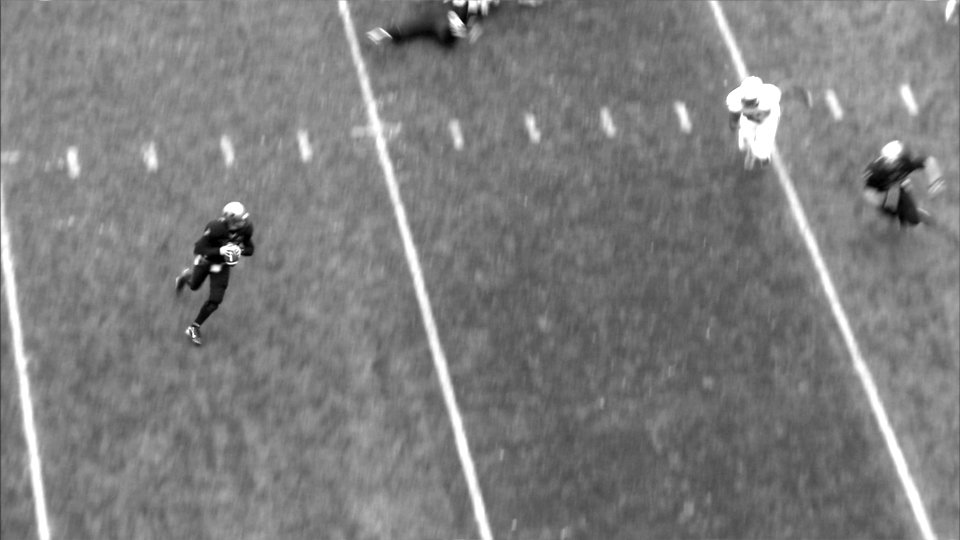}%
    
	 \caption{From left to right: result of VBM3D (our implementation with
		default parameters), VBM3D ST, VBM3D OF, VBM3D ST+OF, VBM3D ST+OF+MS (Lanczos multiscaler) and
	 	the original clean video. The noise level is $\sigma=40$. Contrast has been
		linearly scaled for better visualization.}
    \label{fig:visual}
\end{figure}
\end{landscape}

The first four columns of Figure \ref{fig:visual} show results with/out motion-compensated search and spatio-temporal patches. 
From a qualitative point of view, using spatio-temporal patches provides better temporal consistency. In addition, the patch distance is more reliable since the number of pixels in the patches is doubled. This help retrieve details and texture for regions with a simple motion (e.g. translational). For low noise levels, the effect of these 3d patches is mixed. While the results seem more  consistent temporally, they are blurrier for sequences with complex motions. This explains the drop in PSNR observed for \emph{pedestrians, sunflower} and \emph{touchdown} between VBM3D and VBM3D ST for $\sigma = 10$.
As the noise level increases this  detail loss is out-weighted by the increased robustness to noise of the patch matching.

When spatio-temporal patches are used in conjunction with the optical-flow-guided search, their positive impact is magnified. Although the 
patches are not themselves motion-compensated, having a motion-compensated search region helps find better matches, even in sequences with more complex motion patterns. The motion-compensated search region also improves the temporal consistency, although to a lesser extent than the spatio-temporal patches. The best result, both in terms of PSNR and temporal consistency, is obtained when both strategies are used together (VBM3D ST+OF).

\begin{table*}[htp!]
	\begin{center}
		{\small
		\renewcommand{\tabcolsep}{1.0mm}
		\renewcommand{\arraystretch}{1.}
		\begin{tabular}{@{} c l c c c c c c c @{\hskip 1cm} c @{}}
		\toprule
            \rule{0pt}{6pt}$\sigma$
            & Method                                                             &    {Crowd} &  {Park} & {Pedestrians} &  {Station} &{Sunflower} &{Touchdown} & {Tractor}  & average   \\\hline
            \multirow{1}{*}{$10$}
            & VBM3D~\cite{Dabov2007v}                                            &    35.65   &    34.75   &    40.83   &    38.93   &    40.49   &    39.04   &    37.01   &    38.10  \\
            & VBM3D (ours)                                                       &    35.52   &    34.59   &    40.65   &    38.38   &    39.81   &    39.01   &    36.82   &    37.83  \\
            & VBM3D ST                                                           &\SB{35.65}  &    34.66   &    40.41   &    38.55   &    39.65   &    38.91   &    36.90   &    37.82  \\
            & VBM3D OF                                                           &    35.61   &\SB{34.94}  & \B{41.04}  &\SB{39.79}  &\SB{41.75}  &\SB{39.89}  &\SB{38.43}  &\SB{38.78} \\
            & VBM3D ST+OF                                                        & \B{35.74}  & \B{35.04}  &\SB{41.01}  & \B{40.41}  & \B{41.91}  & \B{39.98}  & \B{38.71}  & \B{38.97} \\
            \midrule
            \multirow{1}{*}{$20$} 
            & VBM3D~\cite{Dabov2007v}                                            &    32.25   &    31.25   &    36.94   &    35.45   &    36.46   &    36.08   &    33.07   &    34.50  \\
            & VBM3D (ours)                                                       &    32.06   &    31.12   &    36.81   &    35.10   &    35.95   &    36.05   &    32.97   &    34.30  \\
            & VBM3D ST                                                           &\SB{32.39}  &    31.36   &    36.97   &    35.57   &    36.14   &    36.16   &    33.23   &    34.55  \\
            & VBM3D OF                                                           &    32.17   &\SB{31.44}  &    37.32   &    36.26   &\SB{38.02}  &\SB{36.91}  &    34.58   &\SB{35.24} \\
            & VBM3D ST+OF                                                        & \B{32.48}  & \B{31.71}  & \B{37.61}  & \B{37.02}  & \B{38.45}  & \B{37.19}  & \B{35.18}  & \B{35.66} \\
            \midrule
            \multirow{1}{*}{$40$} 
            & VBM3D~\cite{Dabov2007v}                                            &    28.65   &    27.68   &    32.81   &    32.02   &    32.65   &    33.52   &    29.41   &    30.96  \\
            & VBM3D (ours)                                                       &    28.39   &    27.64   &    32.62   &    31.80   &    32.31   &    33.35   &    29.38   &    30.78   \\
            & VBM3D ST                                                           &\SB{29.18}  &    28.13   &    33.35   &    32.50   &    32.70   &    33.65   &    29.66   &    31.31  \\
            & VBM3D OF                                                           &    28.55   &    27.99   &    33.29   &    32.80   &\SB{34.46}  &    34.00   &    30.79   &    31.70  \\
            & VBM3D ST+OF                                                        & \B{29.30}  & \B{28.50}  &\SB{34.21}  & \B{33.68}  & \B{35.06}  & \B{34.47}  & \B{31.46}  & \B{32.38} \\
			\bottomrule
		\end{tabular}}
	\end{center}
    \caption{Quantitative denoising results (PSNR and SSIM) for seven grayscale test
		sequences of size $960\times 540$ from the \textit{Derf's Test Media
collection} for several variants of VBM3D. We highlighted the
best performance in black and the second best in brown.}
	\label{tab:of-st-psnr-classic-gray}
\end{table*}

\subsection{Multiscale video denoising}
\label{sec:multiscale}

Multiscale approaches have shown to both reduce the residual noise but also improve the visual quality of the result. 
Indeed, most denoising algorithms work by processing local regions of the image/video (a patch, a group of patches, the receptive field of a neuron, etc). As a result, these methods fail to remove the lower frequencies of the noise. This results in smooth bumps mostly noticeable in large non-textured areas.
Multiscale approaches are able to reduce this artifact by applying the denoising algorithm at different scales. 

There are two main approaches for multi-scale denoising in the literature. The first one consists in modifying the denoising algorithm to consider several scales of the image/video. See for instance the multiscale version of the EPLL method \cite{Zoran2011} proposed by Papyan and Elad \cite{papyan2016multi}. Another approach proposed in \cite{facciolo-multiscaler-siims-17,ipol.2017.201} considers the denoising algorithm as a black box, applying it as is at multiple scales. The result is then obtained by merging the results of each scale. This has the benefit that it can be applied to any denoising method, without any adaptation needed.

The multiscaler of \cite{facciolo-multiscaler-siims-17} first creates a pyramid from the noisy input image $v^0 = v$ by applying a downscaling operator $\mathcal D$
\begin{equation}
v^{s+1} = \mathcal D(v^s), \text{ for } s = 0,...,S-1.
\end{equation}
At each scale, the denoising algorithm is applied and yields  denoised images $\hat u^{s}$. These are then recomposed into a single multiscale output $\hat u_{\text{ms}}$. The recomposition  is recursive. The recursion is initialized at the coarsest level by defining $\hat u_{\text{ms}}^{S} = \hat u^S$, and then proceeds as follows:
\begin{equation}
\hat u_{\text{ms}}^s = \hat u^s - \mathcal U(\mathcal L(\mathcal D(\hat u^s), f_{\text{rec}})) + \mathcal U(\mathcal L(\hat u_{\text{ms}}^{s+1}, f_{\text{rec}})),
\label{eq:ms-recomposition}
\end{equation}
where $U$ is the upscaling operator, and $\mathcal L(\,\cdot\,,f_{\text{rec}})$ is a low-pass filter with cutoff frequency parameterized by $f_{\text{rec}}.$
This equation substitutes the low frequencies of the single-scale result $\hat u^s$ with the multiscale solution $\hat u_{\text{ms}}^{s+1}$, computed using the coarser scales $s+1,...,S$. The recomposition parameter $f_{\text{rec}}$ determines which low frequencies are substituted. In one extreme, the low-pass filter lets all frequencies pass, in which case the whole 
coarse solution is used. At the other end, $f_{\text{rec}}$ filters out all frequencies: the solution at the coarser level $\hat u_{\text{ms}}^{s+1}$ is 
discarded, and the output of the recomposition is the single scale denoising $\hat u^0$. In \cite{facciolo-multiscaler-siims-17} it is found that the optimum is to filter out some of the high frequencies of the coarser level $\hat u_{\text{ms}}^{s+1}$. However, the exact amount depends on the denoiser.

To apply the multiscaler on a video, we apply it spatially, \textit{i.e.} the temporal dimension is not downscaled. We first create a spatial pyramid of the entire video by creating a pyramid of each frame. We then denoise these videos, and recompose them by applying Eq. \eqref{eq:ms-recomposition} to each frame.
This is summarized in Algorithm \ref{alg:multiscale}.

\begin{algorithm}
	\caption{$ $: Multi-scale processing}
	\label{alg:multiscale}
	\setstretch{1.4}
	\DontPrintSemicolon
	\Input{A video $v$, a list of scales $S$, parameters for VBM3D, $p$}
	\Output{The denoised video $\hat{v}$}
	\For{each scale $s$ in $S$}{
	    \tcp{Compute the video $v^s$ at the given scale}
	    \For{each frame $v_i$ in $v$}{
	        $v_i^s \leftarrow v_i$ at scale $s$\; 
	    } 
	    \tcp{Denoise the video $v^s$}
	    $\hat{v}^s \leftarrow \text{VBM3D}(v^s, p)$\;
	}
	\For{each frame index $i$}{
	    $\hat{v}_i \leftarrow$ Combine the $v_i^s$ for $s$ in $S$\;
	}
	\Return $\hat{v}$\;
\end{algorithm}

The parameters of the multiscaler are the number of scales, the downsampling ratio and the recomposition parameter $f_{\text{rec}}$. We shall set the downscaling ratio to 2 (the default), and try different values for the number of scales and the recomposition factor.
There are different possibilities for the down/upscaling operators and the low-pass filter. 
\begin{description}
\item[DCT pyramid.] The DCT multiscaler uses a DCT
pyramid which guarantees white Gaussian noise at all scales. The downscaling is performed by computing the DCT transform of the image and keeping only the quadrant of the image corresponding to the lowest frequencies. The upscaling is done by zero-padding in the DCT domain, and the low-pass filtering zeroes out the highest frequencies in the DCT domain. In this case $f_{\text{rec}}$ represents the ratio of frequencies that are left: $f_{\text{rec}} = 1$ corresponds to an all-pass filter, where as $f_{\text{rec}} = 0$ filters out all the image.
\item[Laplacian pyramid.] The downscaling and upscaling operators samples the image using a Lanczos kernel:
\begin{equation}
    k_{a}(x) = 
    \left\{
    \begin{array}{l l}
    \text{sinc}(x)\text{sinc}(x/a) & \text{if } |x| < a \\ 
    0 & \text{otherwise.}
    \end{array}
    \right.
\end{equation}
We set $a = 3$. For the downscaling, to reduce aliasing, the image is downsampled using a scaled version of the kernel $k_3(\,\cdot\,/2)$ (as described in \cite{schumacher:filtering}).\footnote{This corresponds to Matlab's \texttt{imresize} scaling function using the \texttt{lanczos3} interpolation kernel.}
The low pass filter used is the Gaussian filter of width $f_{\text{rec}}$. When $f_{\text{rec}}\to\infty$ we obtain the single scale output, and if $f_{\text{rec}} = 0$ no frequencies from the coarser solutions are discarded.
\end{description}

\begin{figure}
    \centering
    \includegraphics[width=0.32\linewidth]{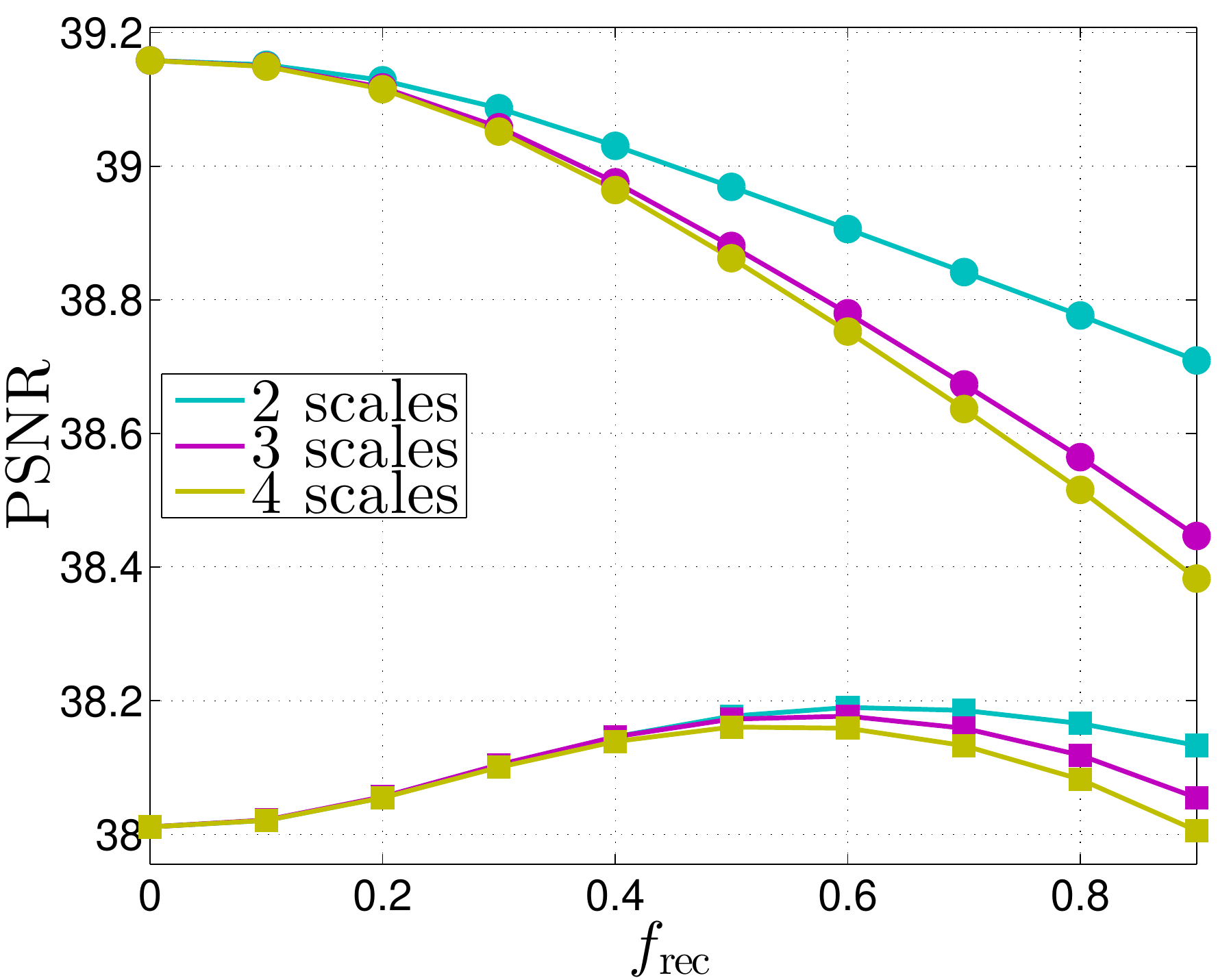}%
    \includegraphics[width=0.32\linewidth]{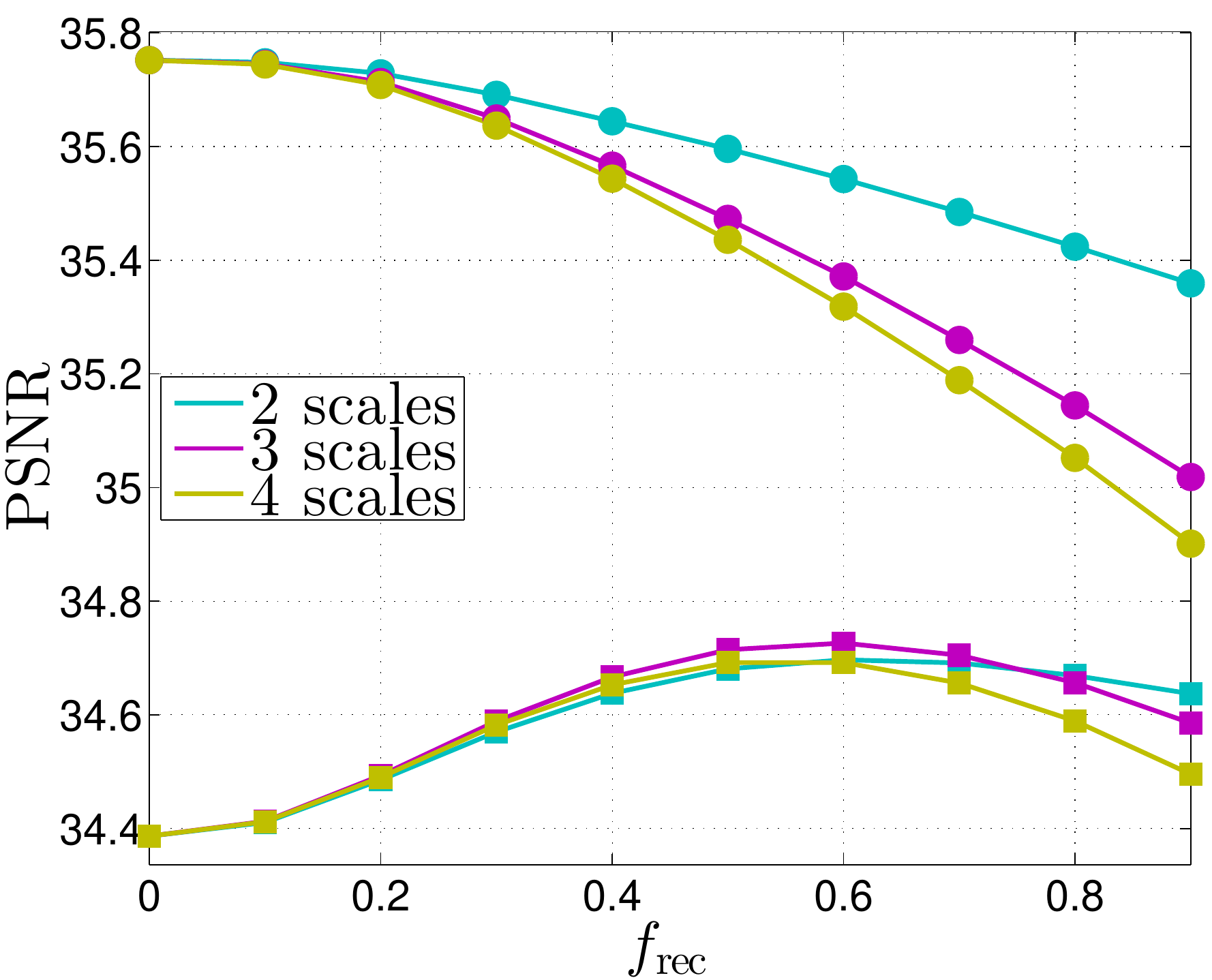}%
    \includegraphics[width=0.32\linewidth]{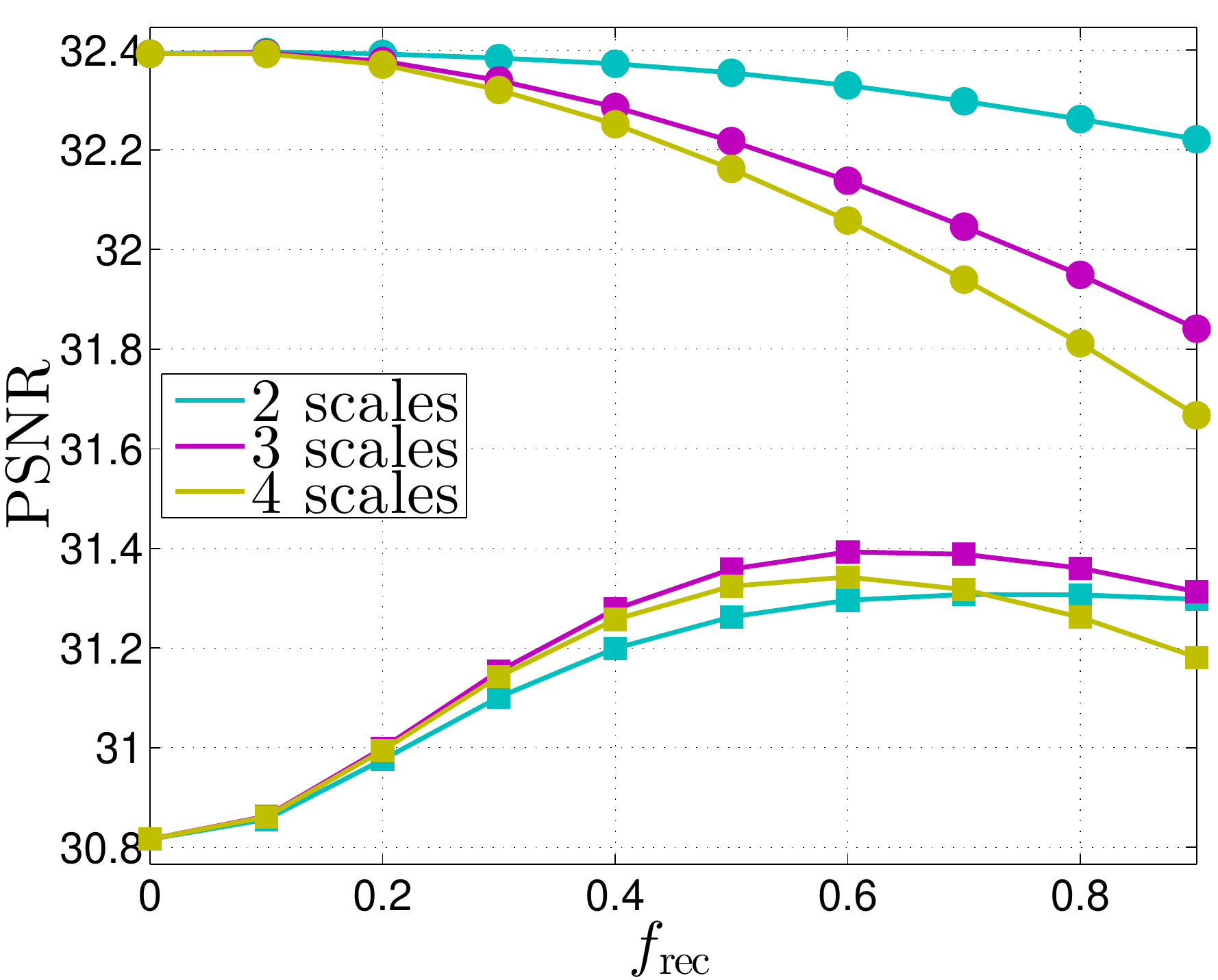}%
    
	 \caption{Effect of the DCT multiscaler for VBM3D without extensions (square markers) and VBM3D ST+OF (round markers). Each plot shows the average PSNR over our seven test sequences obtained when varying the recomposition factor $f_{\text{rec}}$ for different number of scales. From left to right, $\sigma = 10, 20, 40$. The multiscaler has a positive effect on the original VBM3D, but not on the improved VBM3D ST+OF.}
    \label{fig:ms-plot-dct}
\end{figure}

\begin{figure}
    \centering
    \includegraphics[width=0.32\linewidth]{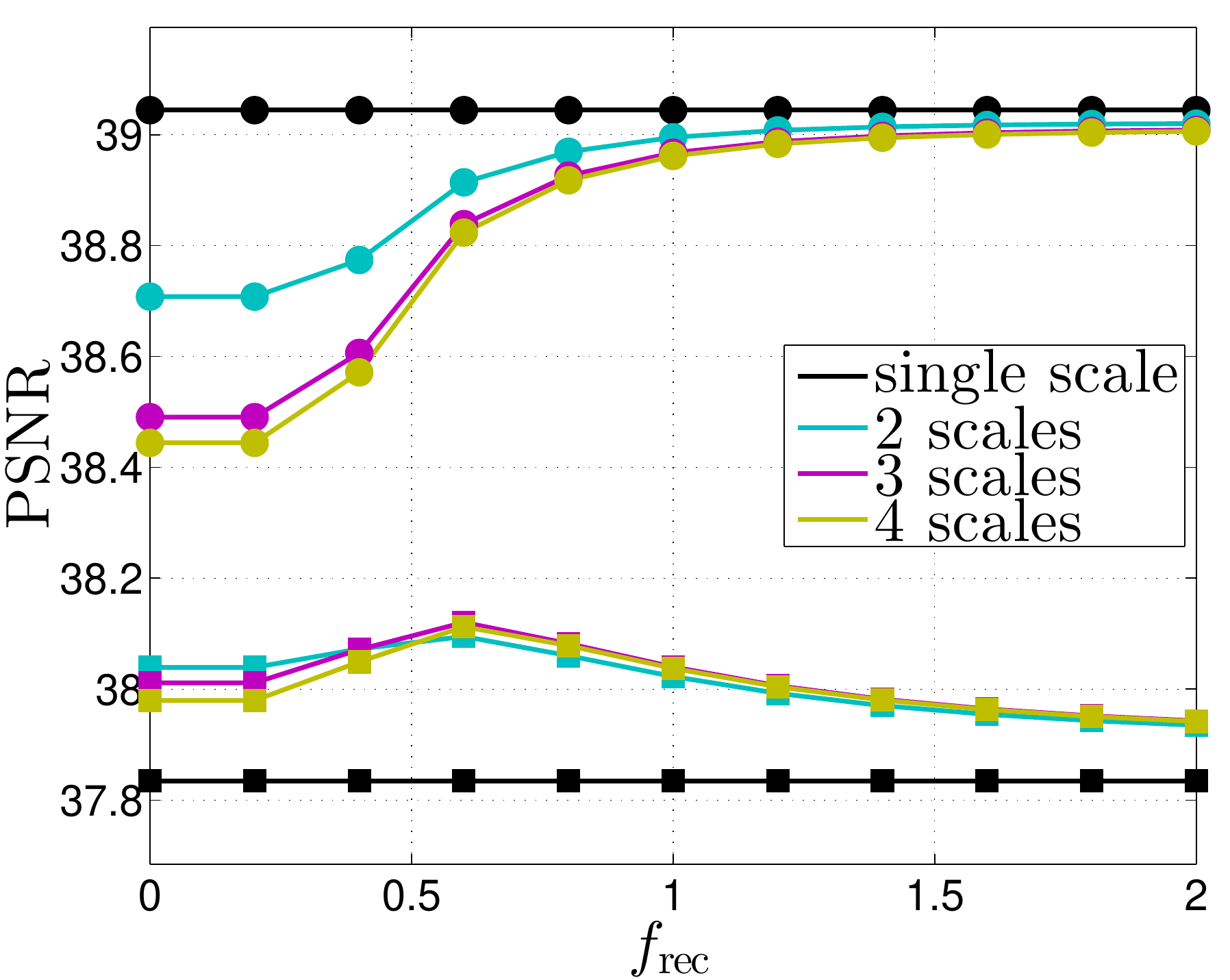}%
    \includegraphics[width=0.32\linewidth]{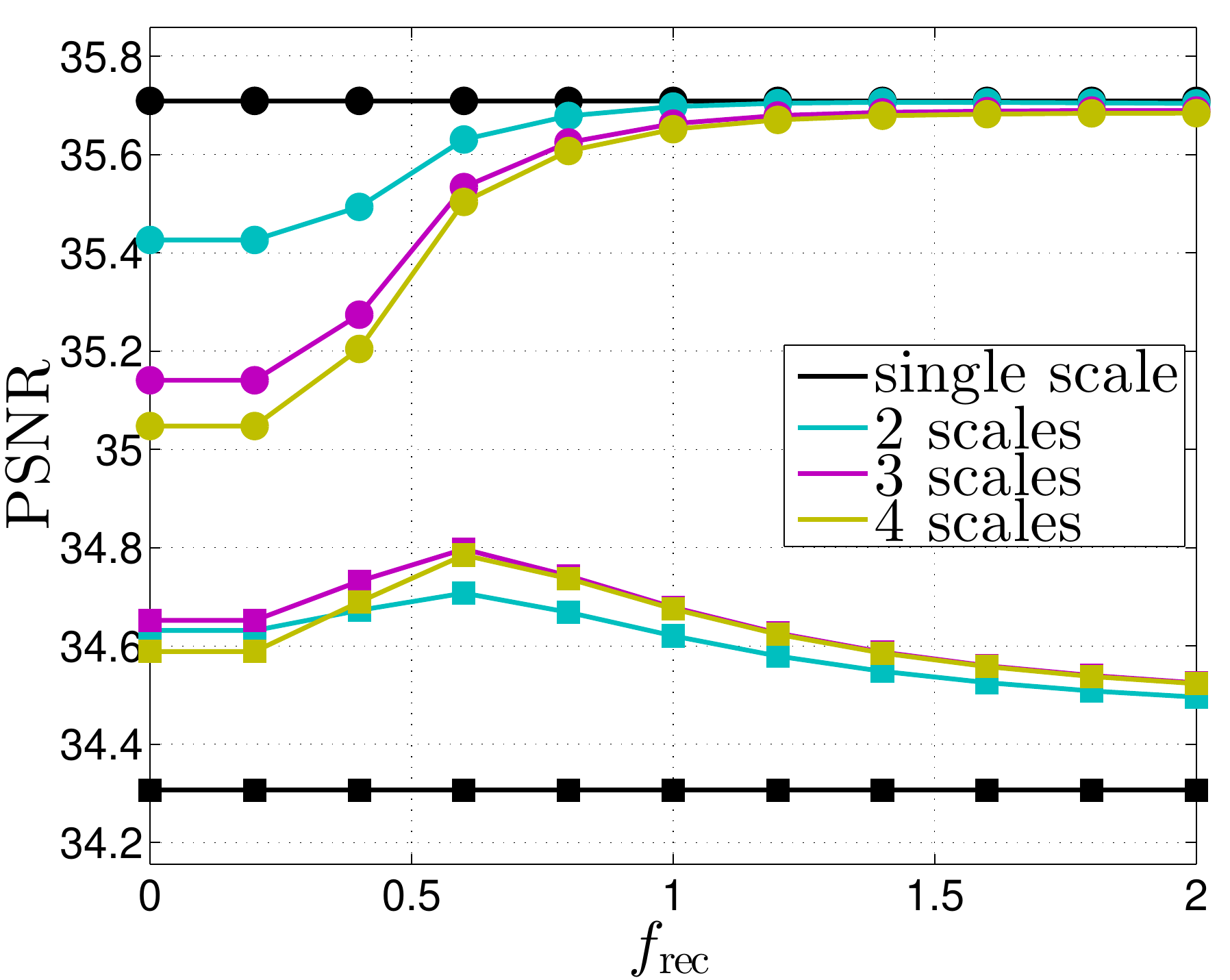}%
    \includegraphics[width=0.32\linewidth]{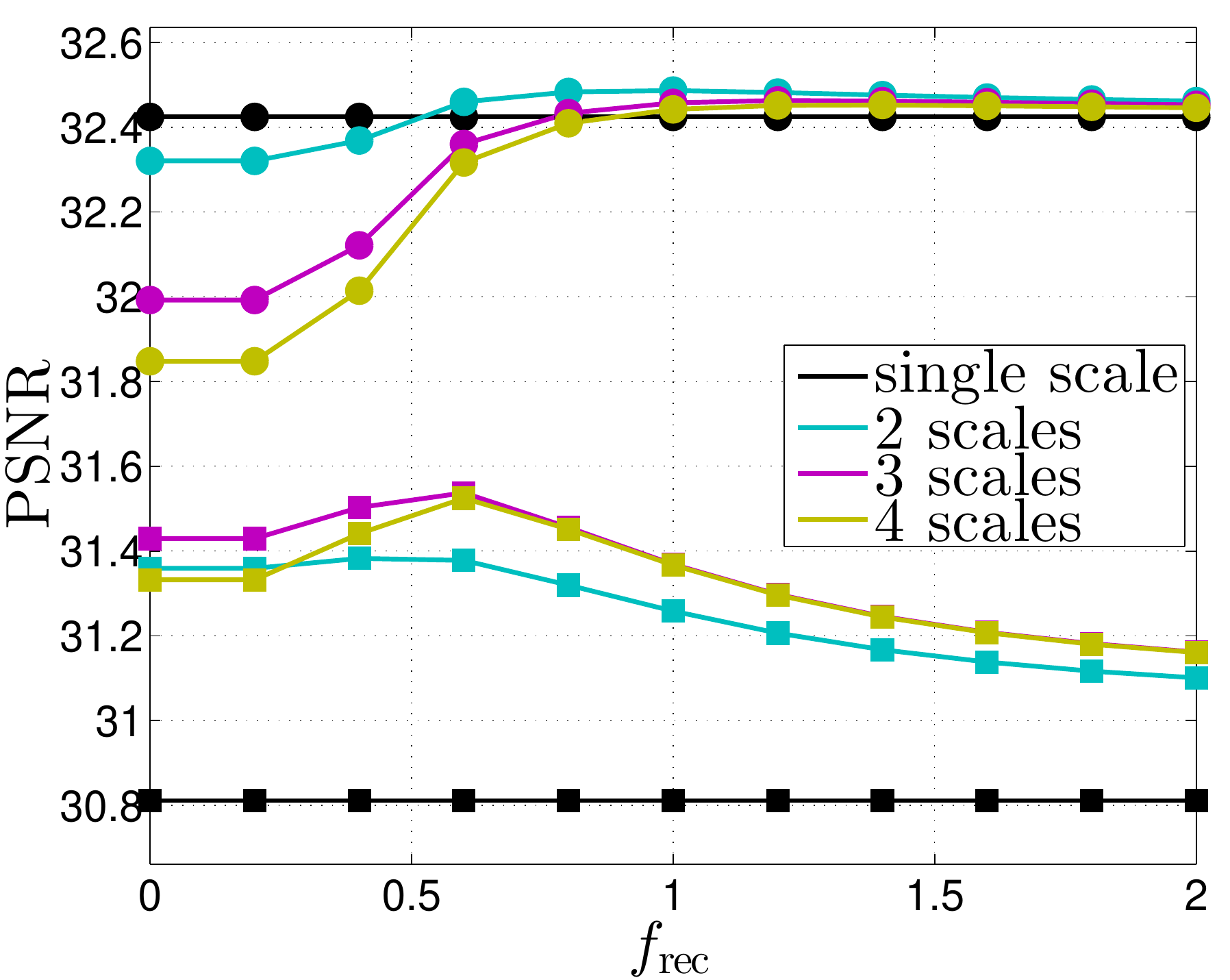}%
    
	 \caption{Effect of the Lanczos multiscaler for VBM3D without extensions
		 (square markers) and VBM3D ST+OF (round markers). Each plot shows the
		 average PSNR over our seven test sequences obtained when varying the
		 recomposition factor $f_{\text{rec}}$ for different number of scales.
	 From left to right, $\sigma = 10, 20, 40$. The multiscaler has a positive
 effect on the original VBM3D, and on the improved VBM3D ST+OF for $\sigma = 40$.}
    \label{fig:ms-plot-lanczos}
\end{figure}

Figures \ref{fig:ms-plot-dct} and \ref{fig:ms-plot-lanczos} show plots of the average PSNR for our seven test sequences obtained with the two multiscalers varying the number of scales and the recomposition cutoff parameter $f_{\text{rec}}$. In each figure, we show the results of the multiscaler applied to the standard VBM3D, and to the one with statio-temporal patches and guided patch search (VBM3D ST+OF). %

Both multiscalers have a positive impact when applied to the standard VBM3D. The gain can be up to 0.3dB for noise 10, 0.5dB for noise 20 and 0.8dB for noise 40. However, when applied to the VBM3D ST+OF version of the algorithm, the multiscaler does not improve the result. In fact, for noise 10 and 20, the best PSNR is attained by the single scale algorithm and the PSNR deteriorates as the cutoff frequency of the low-pass filter is increased (i.e. as more frequency components from the coarse solution are used).  

The multiscaler achieves a better denoising of large objects with smooth textures by removing low-frequency noise left by the denoiser. This low-frequency noise is much stronger for the VBM3D denoiser than for the VBM3D ST+OF version. Hence, the improvement provided by the multiscaler is smaller for the latter. This also depends on the characteristics of the sequence. Frames with smooth objects occupying larger areas will  benefit from the multiscaler.
Yet, the multiscaler introduces artifacts penalizing the PSNR (particularly additional ringing for the DCT multiscaler). These artifacts are not temporally coherent and can therefore be quite noticeable.

Based on these plots we chose to select the Lanczos3 multiscaler. We use 2 scales and a recomposition factor $f_{\text{rec}} = 1$ when applied to VBM3D ST+OF (i.e. VBM3D ST+OF+MS) and 3 scales with recomposition factor $f_{\text{rec}} = 0.6$ when applied to the standard VBM3D (VBM3D MS). Table \ref{tab:ms-psnr-classic-gray} shows the obtained PSNRs. The visual results of VBM3D ST+OF+MS are shown in Figure \ref{fig:visual}. The impact of the multiscaler can be noticed in the top row (results for \textit{pedestrian}) as a reduction of the low-frequency noise in the smooth areas in the image. For the other sequences, since they are highly textured, the effect of the multiscaler is subtle. A careful  examination reveals some texture loss.

\begin{table*}[htp!]
	\begin{center}
		{\small
		\renewcommand{\tabcolsep}{1.0mm}
		\renewcommand{\arraystretch}{1.}
		\begin{tabular}{@{} c l c c c c c c c @{\hskip 0.3cm} c @{}}
		\toprule
            \rule{0pt}{6pt}$\sigma$
            & Method                                                             &    {Crowd} &  {Park} & {Pedestrians} &  {Station} &{Sunflower} &{Touchdown} & {Tractor}  & average   \\\hline
            \multirow{1}{*}{$10$}
            & VBM3D (ours)                                                       &    35.52   &    34.59   &    40.65   &    38.38   &    39.81   &    39.01   &    36.82   &    37.83  \\
            & VBM3D MS (DCT)                                                     &    35.28   &    34.42   &    40.73   &    38.62   &    40.51   &    39.23   &    36.93   &    37.96  \\
            & VBM3D MS (Lanczos)                                                 &    35.48   &    34.57   &    40.79   &    38.59   &    40.35   &    39.19   &    36.93   &    37.99  \\
            & VBM3D ST+OF                                                        & \B{35.74}  & \B{35.04}  &\SB{41.01}  & \B{40.41}  & \B{41.91}  & \B{39.98}  & \B{38.71}  & \B{38.97} \\
            & VBM3D ST+OF+MS (DCT)                                               &    35.48   &    34.74   &    40.82   &    39.56   &    41.16   &    39.62   &    38.06   &    38.49  \\
            & VBM3D ST+OF+MS (Lanczos)                                           &\SB{35.72}  &\SB{35.01}  & \B{41.05}  &\SB{40.34}  &\SB{41.88}  &\SB{39.97}  &\SB{38.66}  &\SB{38.95} \\
            \midrule
            \multirow{1}{*}{$20$} 
            & VBM3D (ours)                                                       &    32.06   &    31.12   &    36.81   &    35.10   &    35.95   &    36.05   &    32.97   &    34.30  \\
            & VBM3D MS (DCT)                                                     &    31.75   &    30.92   &    37.31   &    35.49   &    37.22   &    36.33   &    33.49   &    34.64  \\
            & VBM3D MS (Lanczos)                                                 &    32.04   &    31.13   &    37.29   &    35.45   &    36.95   &    36.31   &    33.35   &    34.65  \\
            & VBM3D ST+OF                                                        & \B{32.48}  & \B{31.71}  &\SB{37.61}  & \B{37.02}  & \B{38.45}  & \B{37.19}  & \B{35.18}  &\SB{35.66} \\
            & VBM3D ST+OF+MS (DCT)                                               &    32.08   &    31.32   &    37.47   &    36.27   &\SB{37.71}  &    36.71   &\SB{34.51}  &    35.15  \\
            & VBM3D ST+OF+MS (Lanczos)                                           &\SB{32.46}  &\SB{31.68}  & \B{37.74}  &\SB{36.99}  & \B{38.45}  &\SB{37.16}  & \B{35.18}  & \B{35.67} \\
            \midrule
            \multirow{1}{*}{$40$} 
            & VBM3D (ours)                                                       &    28.39   &    27.64   &    32.62   &    31.80   &    32.31   &    33.35   &    29.38   &    30.78   \\
            & VBM3D MS (DCT)                                                     &    28.31   &    27.68   &    33.75   &    32.42   &    33.90   &    33.60   &    30.27   &    31.42  \\
            & VBM3D MS (Lanczos)                                                 &    28.51   &    27.78   &    33.54   &    32.31   &    33.55   &    33.64   &    30.01   &    31.33  \\
            & VBM3D ST+OF                                                        & \B{29.30}  &\SB{28.50}  &    34.21   &\SB{33.68}  & \B{35.06}  &\SB{34.47}  &\SB{31.46}  &\SB{32.38} \\
            & VBM3D ST+OF+MS (DCT)                                               &\SB{28.90}  &    28.18   &\SB{34.25}  &    33.19   &\SB{34.41}  &    34.04   &    31.00   &    32.03  \\
            & VBM3D ST+OF+MS (Lanczos)                                           & \B{29.30}  & \B{28.51}  & \B{34.46}  & \B{33.70}  & \B{35.06}  & \B{34.50}  & \B{31.56}  & \B{32.44} \\
			\bottomrule
		\end{tabular}}
	\end{center}
    \caption{Quantitative denoising results (PSNR and SSIM) for seven grayscale test
		sequences of size $960\times 540$ from the \textit{Derf's Test Media
collection} for several variants of VBM3D. We highlighted the
best performance in black and the second best in brown.}
	\label{tab:ms-psnr-classic-gray}
\end{table*}

\section{Comparison with the state of the art}

In Table~\ref{tab:psnr-classic-gray}, we compare the PSNR of different recent denoising methods with VBM3D as well as with versions of VBM3D modified using different combinations of the improvements suggested in Sections \ref{sec:multiscale}, \ref{sec:spatiotemp}, \ref{sec:guided}. VBM3D combined with spatio-temporal patches and a patch-search guided with an optical flow gives very competitive results even compared to the latest state of the art, and especially for higher noises. When combined with the other improvements, it seems that multiscaling does not increase the quality of the denoising and can even degrade it in terms of PSNR. Visual examples are shown in Figures \ref{fig:visual}, where we compare results obtained with the original VBM3D with our implementation, using 3D patches and  an optical flow to guide the search region.

\begin{table*}[htp!]
	\begin{center}
		{\small
		\renewcommand{\tabcolsep}{1.0mm}
		\renewcommand{\arraystretch}{1.}
		\begin{tabular}{@{} c l c c c c c c c @{\hskip 1cm} c @{}}
		\toprule
            \rule{0pt}{6pt}$\sigma$
            & Method                                   &    {Crowd} &  {Park} & {Pedestrians} &  {Station} &{Sunflower} &{Touchdown} & {Tractor}  & average   \\\hline
            \multirow{1}{*}{$10$}
            & VBM3D~\cite{Dabov2007v}                  &    35.65   &    34.75   &    40.83   &    38.93   &    40.49   &    39.04   &    37.01   &    38.10  \\
			& BM4D~\cite{Maggioni2013}                 &    35.84   &    34.45   &    41.15   &    40.23   &    40.97   &    39.78   &    37.33   &    38.54  \\
            & VBM4D~\cite{Maggioni2012}                &    36.05   &    35.31   &    40.61   &    40.85   &    41.88   &    39.79   &    37.73   &    38.88  \\
            & SPTWO~\cite{buades2016patch}             &    36.57   &    35.87   &    41.02   &    41.24   &    42.84   &    40.45   &\SB{38.92}  &    39.56  \\
            & VNLnet~\cite{davy2018non}                &\SB{37.00}  &\SB{36.39}  &\SB{41.96}  & \B{42.44}  & \B{43.76}  &\SB{41.05}  &    38.89   &\SB{40.21} \\
            & VNLB~\cite{Arias2018}                    & \B{37.24}  & \B{36.48}  & \B{42.23}  &\SB{42.14}  &\SB{43.70}  & \B{41.23}  & \B{40.20}  & \B{40.57} \\
            & VBM3D ST+OF+MS                           &    35.72   &    35.01   &    41.05   &    40.34   &    41.88   &    39.97   &    38.66   &    38.95  \\
            \midrule
            \multirow{1}{*}{$20$} 
            & VBM3D~\cite{Dabov2007v}                  &    32.25   &    31.25   &    36.94   &    35.45   &    36.46   &    36.08   &    33.07   &    34.50  \\
            & BM4D~\cite{Maggioni2013}                 &    32.37   &    30.96   &    37.43   &    36.71   &    37.13   &    36.54   &    33.53   &    34.95  \\
            & VBM4D~\cite{Maggioni2012}                &    32.40   &    31.60   &    36.72   &    36.84   &    37.78   &    36.44   &    33.95   &    35.10  \\
            & SPTWO~\cite{buades2016patch}             &    32.94   &    32.35   &    37.01   &    38.09   &    38.83   & \B{37.55}  &    35.15   &    35.99  \\
            & VNLnet~\cite{davy2018non}                &\SB{33.40}  & \B{32.94}  &\SB{38.32}  &\SB{38.49}  & \B{39.88}  &    37.11   &\SB{35.23}  &\SB{36.47} \\
            & VNLB~\cite{Arias2018}                    & \B{33.49}  &\SB{32.80}  & \B{38.61}  & \B{38.78}  &\SB{39.82}  &\SB{37.47}  & \B{36.67}  & \B{36.81} \\
            & VBM3D ST+OF+MS                           &    32.46   &    31.68   &    37.74   &    36.99   &    38.45   &    37.16   &    35.18   &    35.67  \\
            \midrule
            \multirow{1}{*}{$40$} 
            & VBM3D~\cite{Dabov2007v}                  &    28.65   &    27.68   &    32.81   &    32.02   &    32.65   &    33.52   &    29.41   &    30.96  \\
            & BM4D~\cite{Maggioni2013}                 &    29.10   &    27.82   &    33.44   &    32.98   &    33.06   &    33.68   &    29.84   &    31.42  \\
            & VBM4D~\cite{Maggioni2012}                &    28.72   &    27.99   &    32.62   &    32.93   &    33.66   &    33.68   &    30.20   &    31.40  \\
            & SPTWO~\cite{buades2016patch}             &    29.02   &    28.79   &    31.32   &    32.37   &    32.61   &    31.80   &    30.61   &    30.93  \\
            & VNLnet~\cite{davy2018non}                &\SB{29.69}  & \B{28.29}  &    34.21   &\SB{33.96}  &\SB{35.12}  &    33.88   &    31.41   &\SB{32.51} \\
            & VNLB~\cite{Arias2018}                    & \B{29.88}  &\SB{29.28}  & \B{34.68}  & \B{34.65}  & \B{35.44}  &\SB{34.18}  & \B{32.58}  & \B{32.95} \\
            & VBM3D ST+OF+MS                           &    29.30   &    28.51   &\SB{34.46}  &    33.70   &    35.06   & \B{34.50}  &\SB{31.56}  &    32.44  \\
			\bottomrule
		\end{tabular}}
	\end{center}
    \caption{Quantitative denoising results (PSNR and SSIM) for seven grayscale test
		sequences of size $960\times 540$ from the \textit{Derf's Test Media
collection} on several state-of-the-art video denoising algorithms. We highlighted the
best performance in black and the second best in brown.}
	\label{tab:psnr-classic-gray}
\end{table*}

\begin{landscape}
\begin{figure}
    \centering
    \includegraphics[width=0.166\linewidth,trim={4cm 7cm 16cm 0cm}, clip]{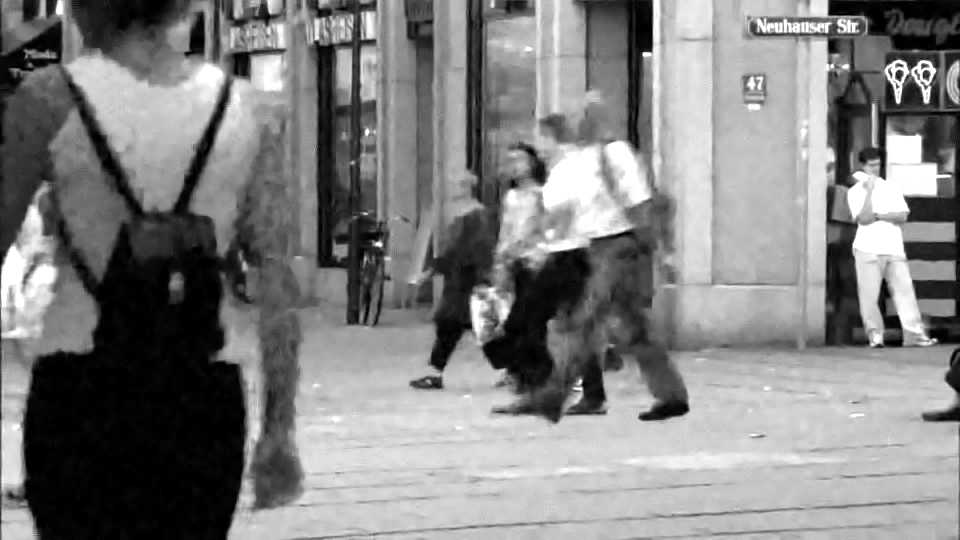}%
    \includegraphics[width=0.166\linewidth,trim={4cm 7cm 16cm 0cm}, clip]{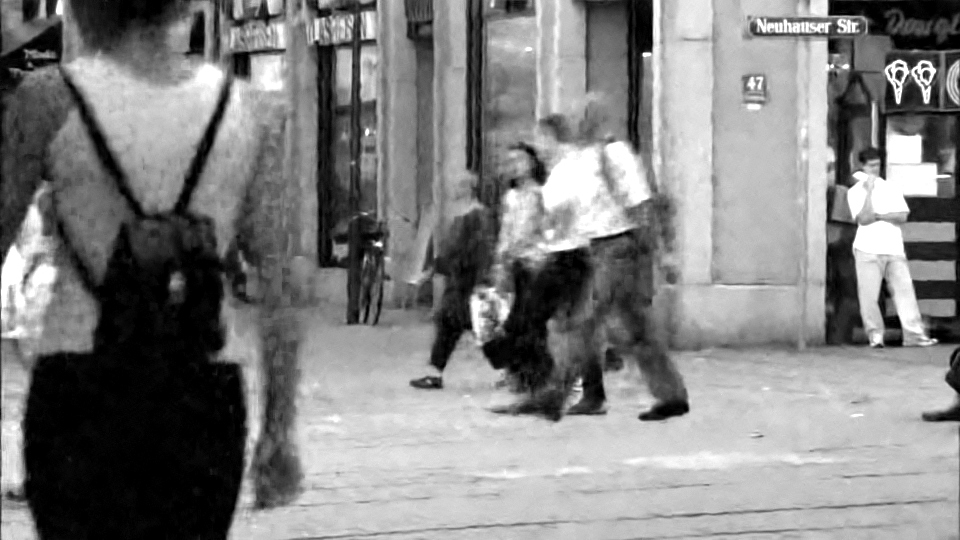}%
    \includegraphics[width=0.166\linewidth,trim={4cm 7cm 16cm 0cm}, clip]{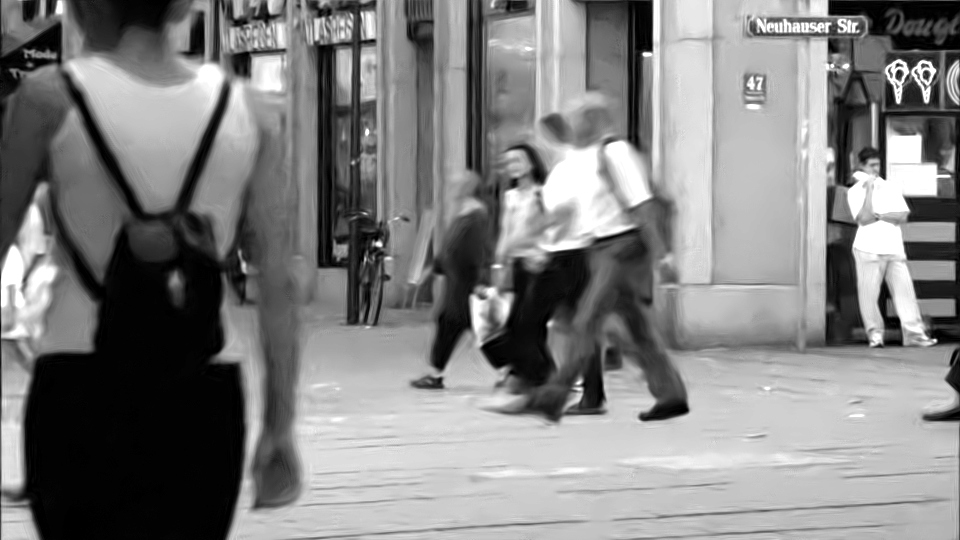}%
    \includegraphics[width=0.166\linewidth,trim={4cm 7cm 16cm 0cm}, clip]{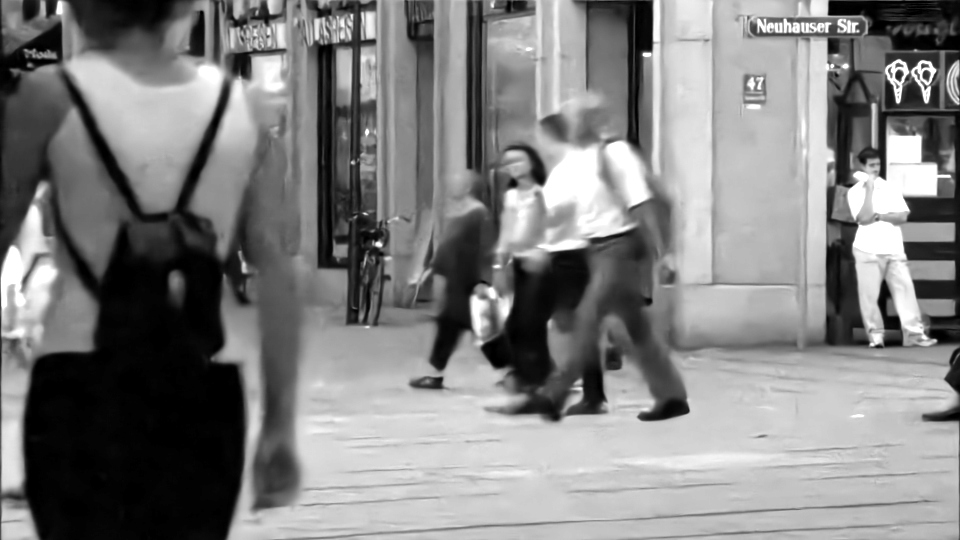}%
    \includegraphics[width=0.166\linewidth,trim={4cm 7cm 16cm 0cm}, clip]{results/0-4-45-345/enhanced-345-lz3-pedestrian_area_055_40.jpg}%
    \includegraphics[width=0.166\linewidth,trim={4cm 7cm 16cm 0cm}, clip]{results/0-4-45-345/enhanced-pedestrian_area_055.jpg}

    \includegraphics[width=0.166\linewidth,trim={16cm 0cm 0cm 4cm}, clip]{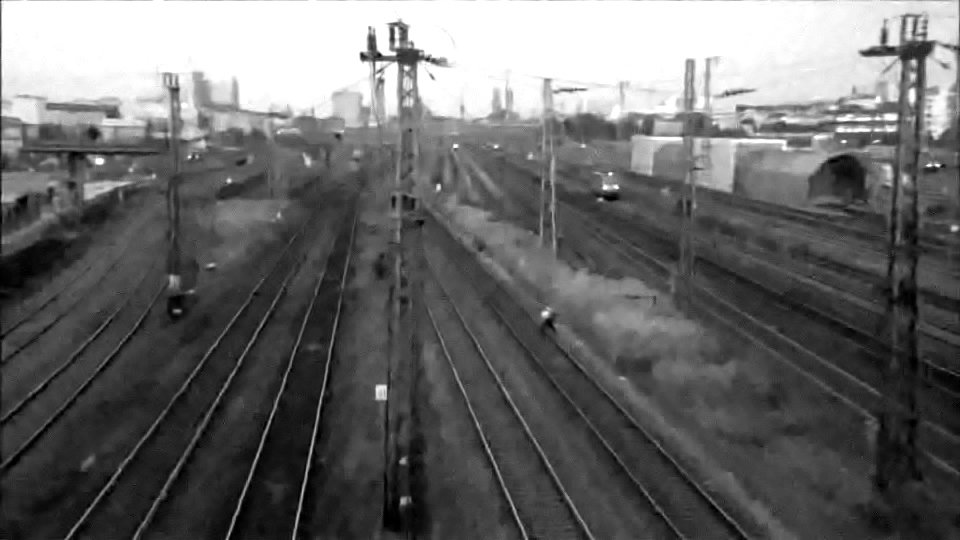}%
    \includegraphics[width=0.166\linewidth,trim={16cm 0cm 0cm 4cm}, clip]{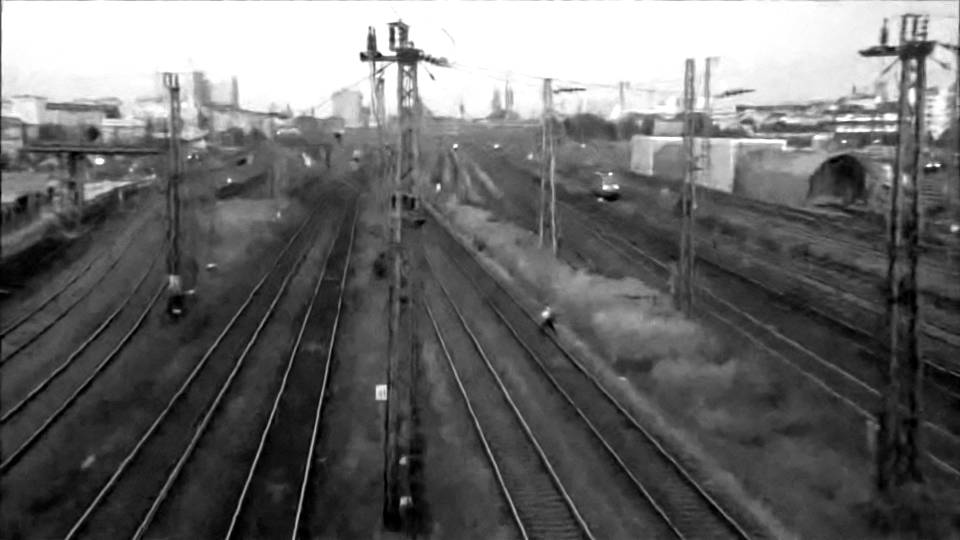}%
    \includegraphics[width=0.166\linewidth,trim={16cm 0cm 0cm 4cm}, clip]{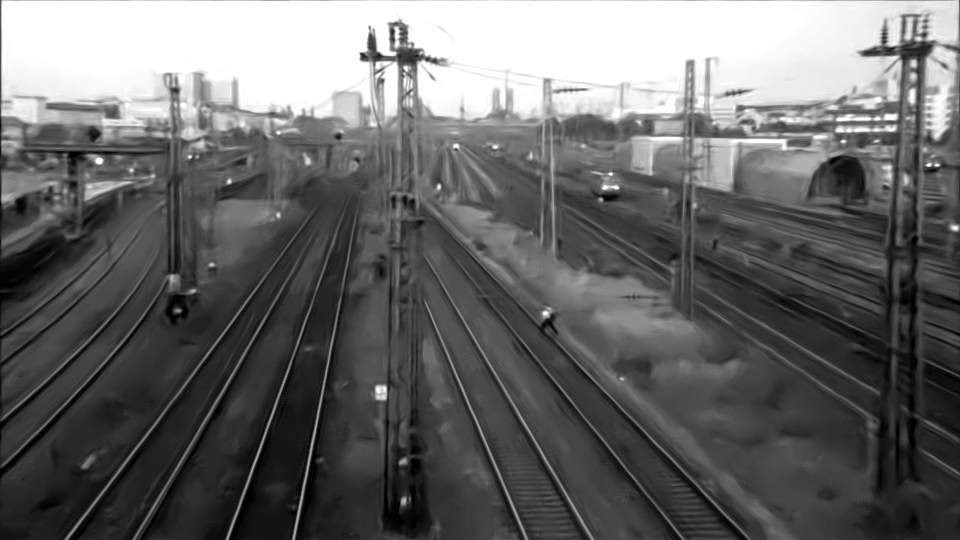}%
    \includegraphics[width=0.166\linewidth,trim={16cm 0cm 0cm 4cm}, clip]{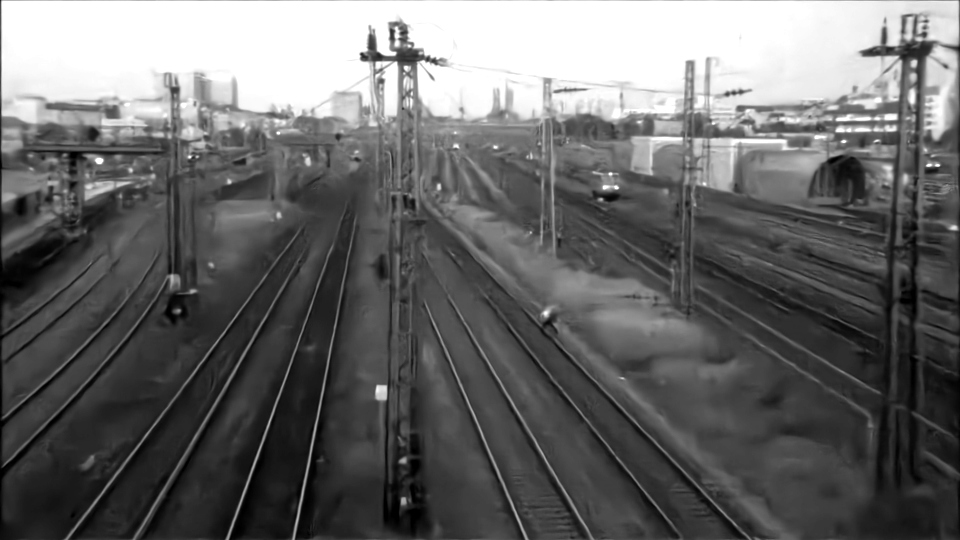}%
    \includegraphics[width=0.166\linewidth,trim={16cm 0cm 0cm 4cm}, clip]{results/0-4-45-345/enhanced-345-lz3-station2_092_40.jpg}%
    \includegraphics[width=0.166\linewidth,trim={16cm 0cm 0cm 4cm}, clip]{results/0-4-45-345/enhanced-station2_092.jpg}%

    \includegraphics[width=0.166\linewidth,trim={1cm 1cm 6cm 0cm}, clip]{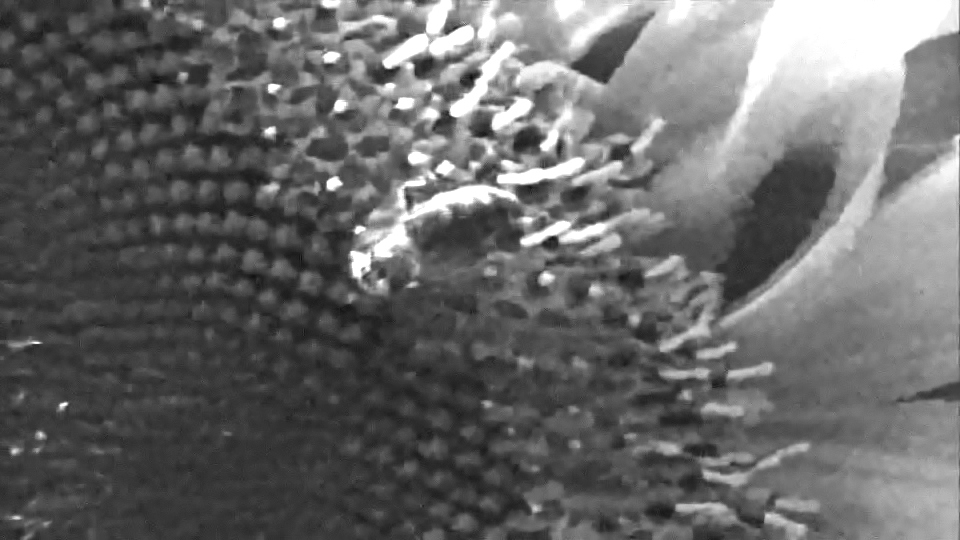}%
    \includegraphics[width=0.166\linewidth,trim={1cm 1cm 6cm 0cm}, clip]{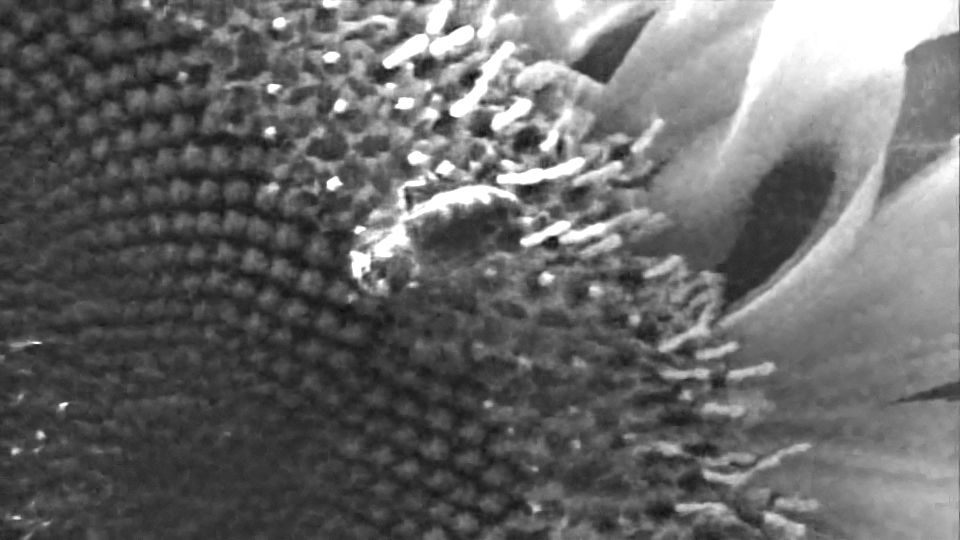}%
    \includegraphics[width=0.166\linewidth,trim={1cm 1cm 6cm 0cm}, clip]{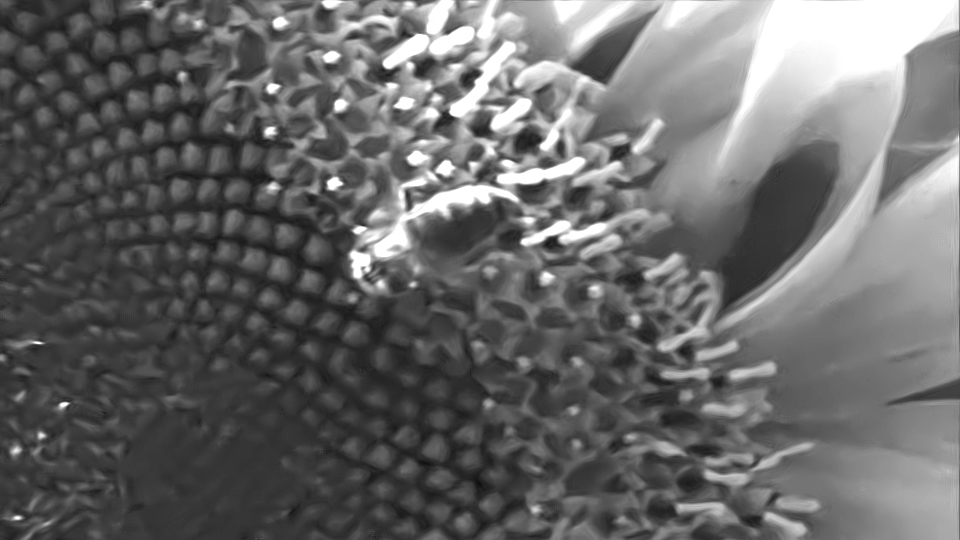}%
    \includegraphics[width=0.166\linewidth,trim={1cm 1cm 6cm 0cm}, clip]{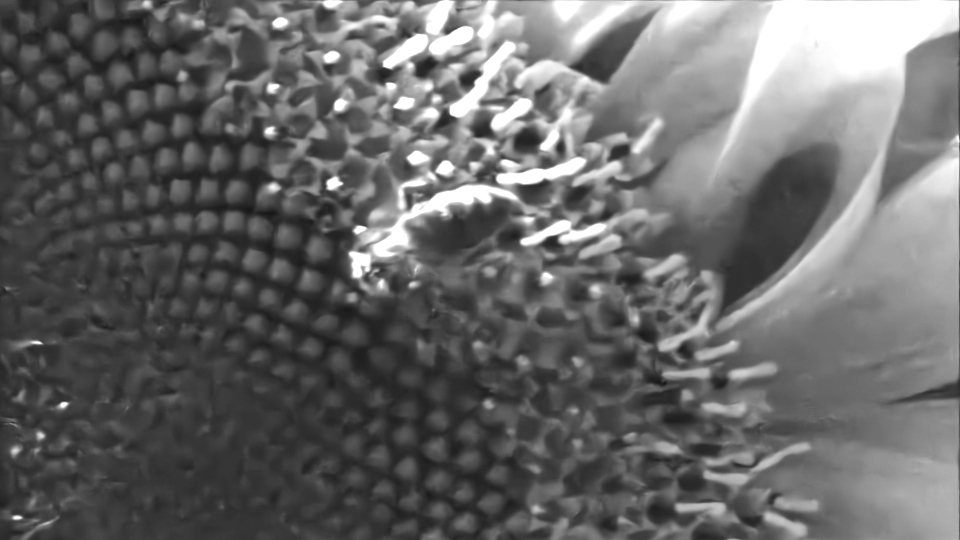}%
    \includegraphics[width=0.166\linewidth,trim={1cm 1cm 6cm 0cm}, clip]{results/0-4-45-345/enhanced-345-lz3-sunflower_022_40.jpg}%
    \includegraphics[width=0.166\linewidth,trim={1cm 1cm 6cm 0cm}, clip]{results/0-4-45-345/enhanced-sunflower_022.jpg}%

    \includegraphics[width=0.166\linewidth,trim={0cm 0cm 16cm 4cm}, clip]{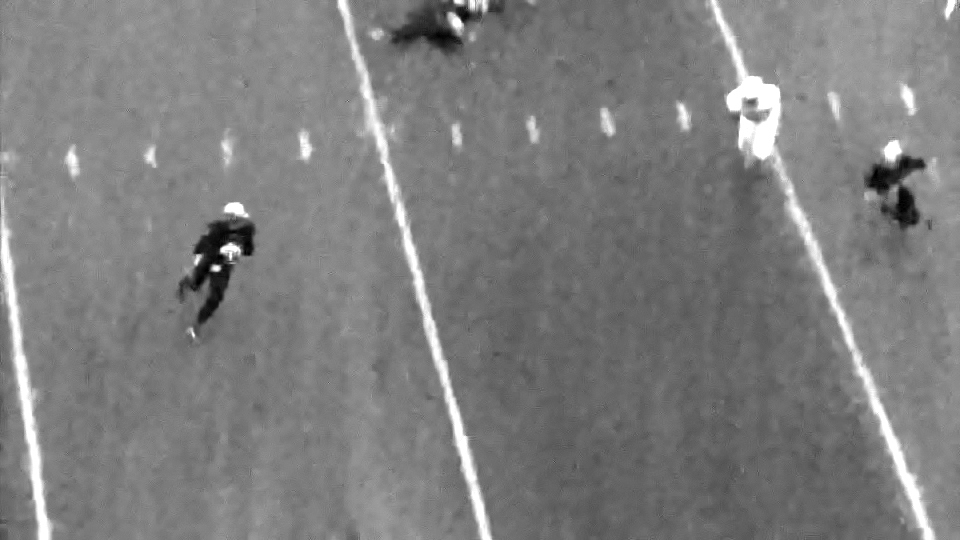}%
    \includegraphics[width=0.166\linewidth,trim={0cm 0cm 16cm 4cm}, clip]{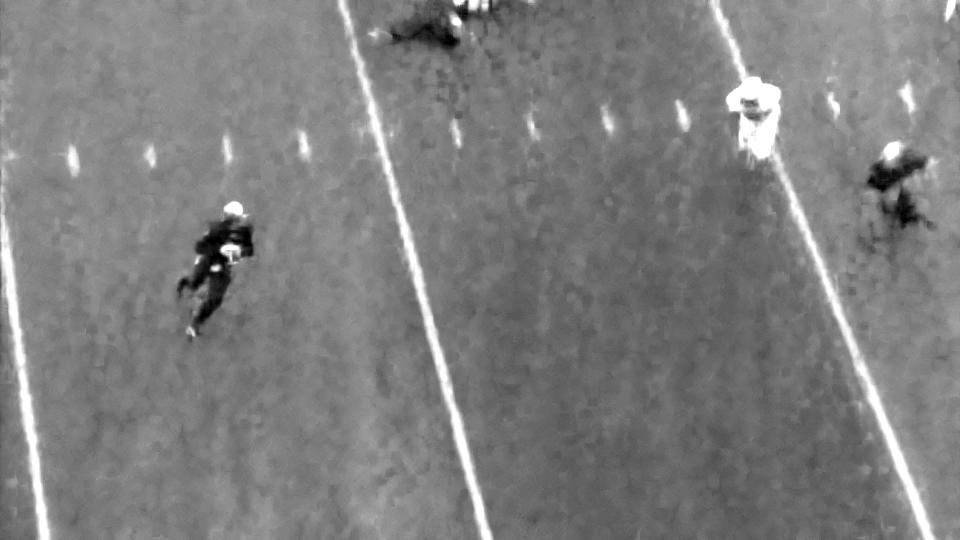}%
    \includegraphics[width=0.166\linewidth,trim={0cm 0cm 16cm 4cm}, clip]{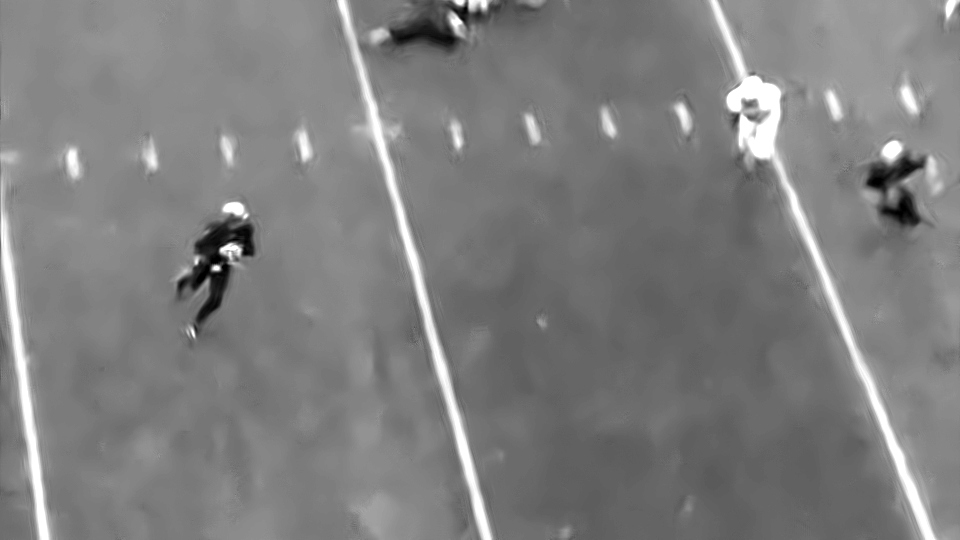}%
    \includegraphics[width=0.166\linewidth,trim={0cm 0cm 16cm 4cm}, clip]{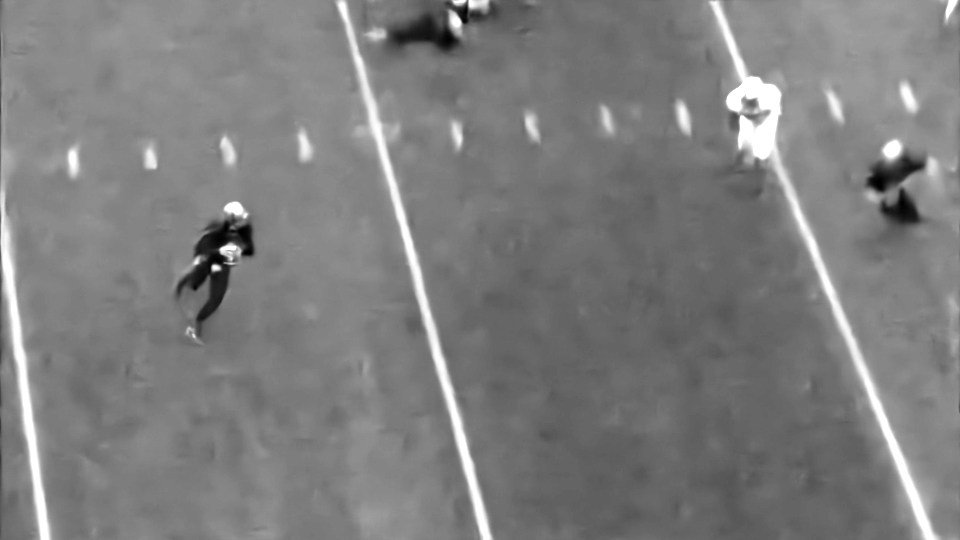}%
    \includegraphics[width=0.166\linewidth,trim={0cm 0cm 16cm 4cm}, clip]{results/0-4-45-345/enhanced-345-lz3-touchdown_pass_182_40.jpg}%
	 \includegraphics[width=0.166\linewidth,trim={0cm 0cm 16cm 4cm}, clip]{results/0-4-45-345/enhanced-touchdown_pass_182.jpg}%

	 \caption{Comparison with state-of-the-art video denoising methods ($\sigma = 40$). From
		 left to right: result of VBM3D \cite{Dabov2007v}, VBM4D
		 \cite{Maggioni2012}, VNLB \cite{Arias2018}, VNLnet \cite{davy2018non},
		 VBM3D SF+OF+MS (Lanczos) and ground truth. Contrast has been linearly scaled for better
		visualization.}
    \label{fig:visual-sota}
\end{figure}
\end{landscape}

{\small
\bibliographystyle{ieee}
\bibliography{image_denoising,video_denoising}
}

\end{document}